\documentclass[11pt]{article}

\usepackage[final]{acl}

\usepackage{times}
\usepackage{latexsym}

\usepackage[T1]{fontenc}

\usepackage[utf8]{inputenc}

\usepackage{microtype}

\usepackage{inconsolata}

\usepackage{graphicx}

\usepackage{siunitx} 
\usepackage[skins]{tcolorbox}
\usepackage{longtable}
\usepackage{threeparttable}
\usepackage{booktabs}  
\usepackage{tabularx} 
\usepackage{caption}  
\usepackage{array}    
\usepackage{wrapfig}
\usepackage{fontawesome}

\usepackage{amsfonts}
\usepackage{amssymb}
\usepackage{amsmath}

\usepackage{enumitem}

\usepackage{subcaption}
\usepackage{graphicx}

\usepackage{longtable,booktabs}

\DeclareSIUnit{\minute}{minute}
\DeclareSIUnit{\hour}{hour}
\DeclareSIUnit{\day}{day}
\DeclareSIUnit{\month}{month}
\DeclareSIUnit{\second}{second}
\title{Your LLM Agents are Temporally Blind:\\ The Misalignment Between Tool Use Decisions and Human Time Perception}

\author{%
  \parbox{0.85\linewidth}{\centering
    Yize Cheng\textsuperscript{1}\thanks{Equal contribution.}~\thanks{Correspondence to Yize Cheng \texttt{<yzcheng@umd.edu>}.},
    Arshia Soltani Moakhar\textsuperscript{1}\footnotemark[1],
    Chenrui Fan\textsuperscript{1}\footnotemark[1],
    Parsa Hosseini\textsuperscript{1},
    Kazem Faghih\textsuperscript{1},
    Zahra Sodagar\textsuperscript{1},
    Wenxiao Wang\textsuperscript{2},
    Soheil Feizi\textsuperscript{1,2}
  }\\[1em]
  \textsuperscript{1}University of Maryland, College Park \quad
  \textsuperscript{2}RELAI.ai\\[0.5em]
  \includegraphics[height=1em]{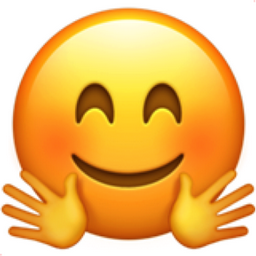}~\textbf{Data}: \url{https://huggingface.co/datasets/yizecheng/TicToc} \\
  \faGithub~\textbf{Code}: \url{https://github.com/chengez/TicToc}
}

\begin{document}
\newcommand{\yize}[1]{\textcolor{cyan}{\scriptsize yize: #1}}
\newcommand{\dsname}{TicToc}
\newcommand{\header}[1]{\noindent\textbf{#1}.}
\maketitle

\begin{abstract}
Large language model (LLM) agents are increasingly used to interact with and execute tasks in dynamic environments. However, a critical yet overlooked limitation of these agents is that they, by default, assume a stationary context, failing to account for the real-world time elapsed between messages. We refer to this as ``temporal blindness''. This limitation hinders decisions about when to invoke tools, leading agents to either over-rely on stale context and skip needed tool calls, or under-rely on it and redundantly repeat tool calls.
To study this challenge, we constructed \mbox{\textbf{\dsname}}, a diverse dataset of multi-turn user–agent message trajectories across 76 scenarios, spanning dynamic environments with high, medium, and low time sensitivity. 
We collected human preferences between ``calling a tool'' and ``directly answering'' on each sample, and evaluated how well LLM tool-calling decisions align with human preferences under varying amounts of elapsed time.
Our analysis reveals that existing models display poor alignment with human temporal perception, with no model achieving a normalized alignment rate better than 65\% when given time stamp information. 
We also show that naive, prompt-based alignment techniques have limited effectiveness for most models, but specific post-training alignment can be a viable way to align multi-turn LLM tool use with human temporal perception. 
Our data and findings provide a first step toward understanding and mitigating temporal blindness, offering insights to foster the development of more time-aware and human-aligned agents.
\end{abstract}
\section{Introduction}

Large language models (LLMs) are increasingly being leveraged as the foundation for agents~\cite{wangvoyager, yao2023react, shinn2023reflexion, wang2023describe}.
These agents are designed to operate in dynamic environments and interact with users to execute complex, real-world tasks by leveraging external tools~\cite{parisi2022talm, schick2023toolformer, song2023restgpt, mialon2023augmented}, such as search engines, and databases. A growing body of research has focused on evaluating how agents use tools, with emphasis on measuring the accuracy of tool invocation~\cite{huang2023metatool, li2023api,patilberkeley}, diagnosing hallucinated calls~\cite{zhang2024toolbehonest,ross2025when2call}, and evaluating the robustness against tool description edits~\cite{shi2025prompt, faghih2025gaming}. However, we highlight that when it comes to tool-use decisions in multi-turn situations, a critical limitation of models' default operational paradigm has been overlooked: Temporal Blindness.


\begin{figure*}[ht]
    \centering
    \includegraphics[width=1\linewidth]{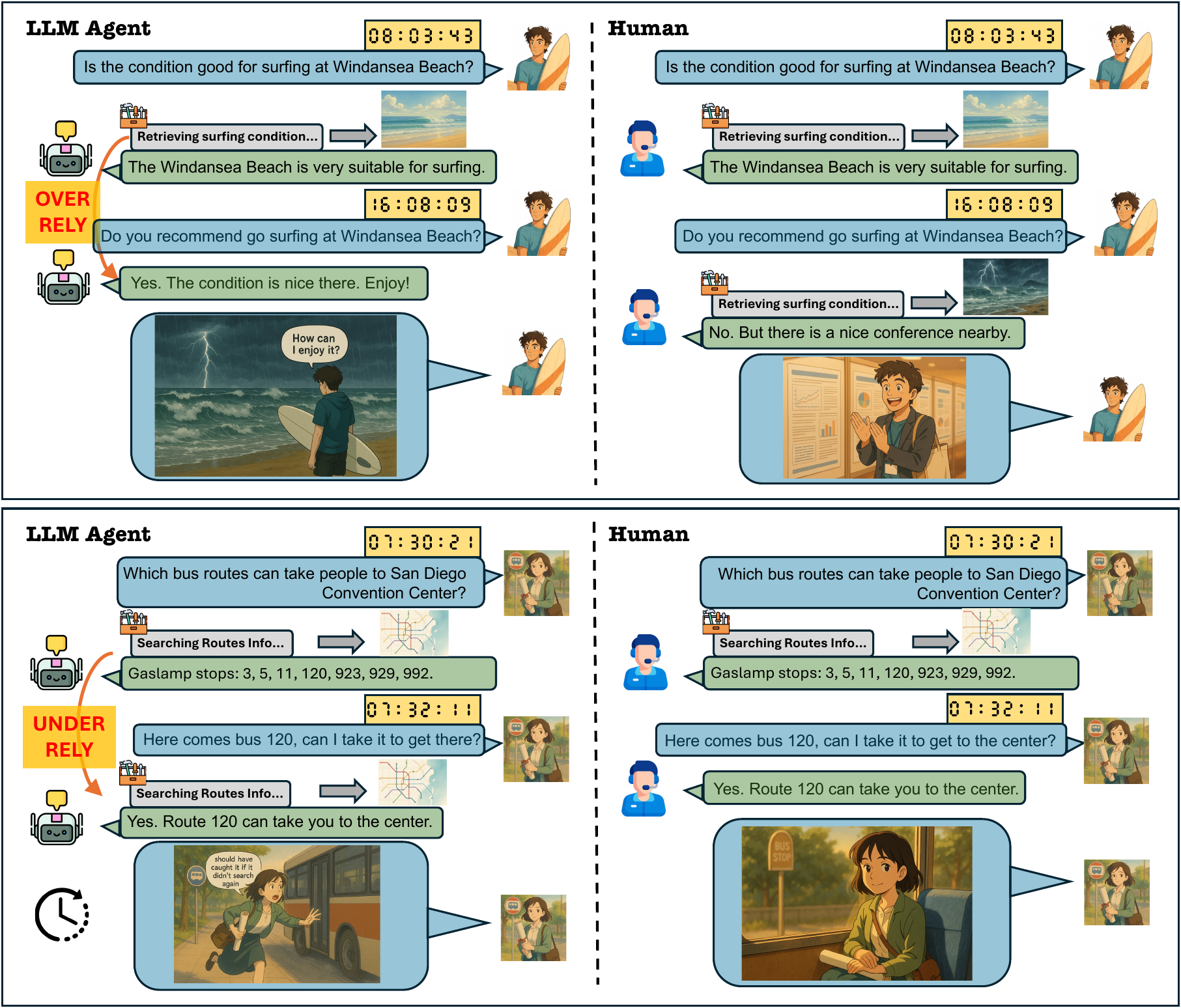}
    \caption{\textbf{Illustrative examples showing the liability of temporal blindness in multi-turn LLM agents, in comparison to human}. The first row shows the case of \underline{over-reliance}, where the model displays over-confidence in outdated context, resulting in erroneous outputs. The second row shows the case of \underline{under-reliance}, where the model displays excessive caution through repeated tool calls, resulting in unnecessary delays and latency.}
    \vspace{-12pt}
    \label{fig:intro}
\end{figure*}

We refer to temporal blindness as the inability of LLM agents to account for the real-world time that elapses between user messages and agent actions. 
While humans naturally integrate the passage of time into their decision-making processes~\cite{Pöppel1978}, LLM agents, by default, operate in environments without intrinsic temporal grounding, making them unable to account for the intervals between messages. Furthermore, our experiments reveal that even when augmented with explicit temporal information, contemporary agents fail to align their tool-use decisions with human preferences regarding elapsed time, underscoring a significant misalignment in their understanding of the external world's dynamics.
Specifically, an agent may either over-rely on a previous observation that is now outdated, thus skipping a necessary tool call, or unnecessarily repeat a tool call for information that is not likely to change (e.g. the radius of the Earth). This, as exemplified in Figure~\ref{fig:intro}, leads to either acting on stale information and producing erroneous outputs in the case of \textbf{over-reliance}, or wasting resources and introducing inefficiency and latency through redundant queries in the case of \textbf{under-reliance}. Neither case aligns with how humans seek help from tools. In practice, humans develop a preference for tool use that is context-sensitive: we generally avoid re-checking stable facts while still knowing when fresh observations are necessary.

To systematically investigate this misalignment, we constructed \textbf{Ti}me-aware \textbf{c}onversational \textbf{To}ol-\textbf{c}alling (\textbf{\dsname}), a diverse dataset with 1800+ multi-turn user-agent message trajectories. \dsname~consists of 76 distinct scenarios, meticulously designed to span environments with high, medium, and low time sensitivity. Each trajectory ends with a user question, where the appropriate reaction to the question, namely whether to rely on prior context or to invoke a tool, is annotated by humans.
We gathered human preferences for each sample by asking annotators whether the agent should call a tool or answer directly. Each sample received at least five annotations, which were then aggregated to determine the final preference. Detailed procedures are provided in Section~\ref{sec:human-preference-collection}.

We evaluate a wide range of contemporary LLM agents on \dsname~under two conditions: with and without explicit timestamp augmentation in the dialogue messages. This setup allows us to measure how temporal information influences tool-calling decisions and their alignment with human preference. Our findings reveal that, in the absence of time signals, agents perform similarly to random guessing in terms of alignment rate. Moreover, even with time information given, the overall alignment rate is still poor, with the best-performing model achieving a normalized alignment rate less than 65\%. We also analyze the correlation between alignment rate and conversation length, and diagnose why reasoning yields little help in improving the alignment results. We further make prompting and post-training based alignment efforts using a subset of \dsname~to offer a first step towards understanding and mitigating the temporal blindness issue in multi-turn LLM agents.
Our core contributions can be summarized as follows:
\begin{itemize}[leftmargin=*]
    \item We identify \textbf{temporal blindness} as a critical limitation of LLM agents in multi-turn interactions, where models fail to account for the passage of real-world time between messages and actions, resulting in either over-reliance or under-reliance on prior context.
    \vspace{-4pt}
    \item We introduce \textbf{Ti}me-aware \textbf{c}onversational \textbf{To}ol-\textbf{c}alling (\textbf{\dsname}), a diverse dataset with 1800+ multi-turn user-agent trajectories across 76 scenarios that vary in time sensitivity, designed to systematically evaluate how well multi-turn LLMs align tool-calling decisions with human temporal perception.
    \vspace{-4pt}
    \item We evaluate a wide range of contemporary LLM agents on \dsname~and perform detailed analysis on their failure modes. We further compare prompting and post-training strategies and show the strong potential of targeted post-training as a necessary step for achieving effective temporal alignment.
\end{itemize}

\section{Related Work}

\subsection{Function calling of LLM agents}

To enable large language model (LLM) agents to be more helpful and extend beyond their parametric knowledge, researchers have introduced function/tool calling capabilities~\cite{parisi2022talm, schick2023toolformer, shen2023hugginggptsolvingaitasks, song2023restgpt, mialon2023augmented}. This has significantly expanded the range of tasks LLMs can perform by enabling interaction with external resources.
Recent standardized interaction protocols, such as the Model Context Protocol (MCP)~\cite{anthropic2024mcp} and Agent2Agent (A2A) Protocol~\cite{google2025a2a}, further enrich the tool ecosystem, streamlining communication and resource access for more sophisticated agentic systems.

Alongside these advancements, a great deal of research has focused on evaluating how LLM agents use tools. Some studies measure the accuracy of tool invocation~\cite{huang2023metatool, li2023api, patilberkeley}, others diagnose hallucinated tool calls~\cite{zhang2024toolbehonest, ross2025when2call}, and a few evaluate robustness to variations in tool descriptions~\cite{shi2025prompt, faghih2025gaming}. However, a significant problem that has been largely overlooked in existing evaluations is temporal blindness in multi-turn settings. Without time awareness, models may either over-rely on outdated context, producing erroneous output, or under-rely on stable context, leading to unnecessary delays from excessive function calls.


\subsection{LLM temporal reasoning}
Temporal reasoning is essential for understanding and interacting with our dynamic world, and has therefore been widely studied in the context of LLMs~\cite{temporal_survey}. Prior work has focused on evaluating~\cite{gupta2023temptabqatemporalquestionanswering, fatemi2025test, wang2024trambenchmarkingtemporalreasoning, chu2024timebenchcomprehensiveevaluationtemporal} and improving~\cite{xiong2024largelanguagemodelslearn, su2024timobettertemporalreasoning, yuan2023futureexplainabletemporalreasoning, TempCoT, liu2025timer1comprehensivetemporalreasoning} LLMs’ ability to understand time-related concepts, order events, and perform temporal deductions. However, these studies largely reason about time in isolation, without situating LLMs in an agentic setting where time continuously evolves and directly affects task execution and decision-making. In contrast, the temporal reasoning capabilities of LLM agents remain underexplored despite their broader practical relevance. Existing agent-oriented work, such as~\citet{ge2025tremuneurosymbolictemporalreasoning}, focuses on time-aware memorization across multi-session dialogues, but overlooks one of the core components of agentic systems—function calling and tool use. Our work instead investigates temporal awareness in LLM agents’ function-calling decisions under multi-turn, time-evolving scenarios, revealing significant misalignment issues arising from agents’ temporal blindness.




\subsection{LLM alignment}
Aligning LLMs with humans is a key post-training challenge~\cite{wang2023aligninglargelanguagemodels,wang2024comprehensivesurveyllmalignment,shen2023alignment}. Pretrained models must be tuned for helpful, trustworthy, and value-aligned behavior using methods such as instruction tuning~\cite{ouyang2022training}, RLHF~\cite{bai2022traininghelpfulharmlessassistant}, and DPO~\cite{rafailov2024directpreferenceoptimizationlanguage}. Alignment spans capabilities like instruction following~\cite{zhou2023instructionfollowingevaluationlargelanguage,qin2024infobenchevaluatinginstructionfollowing}, harm avoidance~\cite{chao2024jailbreakingblackboxlarge}, bias mitigation~\cite{religious_debias,lucy-bamman-2021-gender}, hallucination reduction~\cite{tonmoy2024comprehensivesurveyhallucinationmitigation}, and misuse prevention~\cite{sandbrink2023artificialintelligencebiologicalmisuse}.
Our work introduces the alignment between LLM tool-use decisions and human time perception as an underexplored dimension of the broader LLM alignment problem. We highlight a substantial gap between the behavior of state-of-the-art LLM agents and human expectations, revealing divergences in how humans and LLMs interpret time spans, urgency, and environmental dynamics.


\section{\dsname: Evaluating multi-turn LLM tool-use alignment with human time perception}
\label{sec:dataset}


We present \dsname, a dataset of diverse multi-turn user–agent conversation trajectories with tool calls. This section details its scenario design, trajectory generation pipeline, filtering process, time stamp addition, and human preference collection and aggregation. Each trajectory ends at the final user's question where the appropriate model response is voted by humans between a tool call and a direct response. The dataset is designed to evaluate model alignment in tool use with respect to human time perception after different elapsed time durations when completing tasks of different time sensitivity.

\subsection{Scenario design}
\label{subsec:scenario_design}
We curated \textbf{76 scenarios} covering a range of environments that differ in their temporal dynamics. To capture varying levels of change over time, we classify scenarios into three categories:

\begin{itemize}[leftmargin=*]
    \vspace{-2pt}
    \item \textbf{Low sensitivity} (29 scenarios): environments that are relatively static, where information changes very slowly or not at all (e.g., regulations, published specifications, archival records).
    \item \textbf{Medium sensitivity} (25 scenarios): environments that change from time to time, but in general not abruptly or rapidly (e.g., time slot and reservation booking, forecast and condition reports).
    \item \textbf{High sensitivity} (22 scenarios): environments that are highly dynamic and can change within seconds or minutes (e.g., stock markets, competitive bidding, real-time monitoring).
\end{itemize}

Each scenario can be either \emph{read-only}, where the agent has access only to retrieval functions, or \emph{read+write}, where the agent can both retrieve information and issue actions that modify the environment. A detailed scenario inventory, including sensitivity levels and read/write specifications, is provided in Table~\ref{tab:scenario_invetory}, Appendix~\ref{append:taxonomy}.

\subsection{Trajectory construction}

To capture a diverse range of temporal follow-up behaviors, we first define specific variants for both the \emph{read-only} and \emph{read+write} settings.

For the \emph{read-only} setting, we define four variants. First, in the \textbf{Repeated ask} variant, the user repeats a request for information that was already retrieved in a prior turn. Second, in the \textbf{Comparison} variant, the user retrieves information about item~A and item~B in separate turns, and subsequently asks for a comparison between the two. Third, in the \textbf{Retrieve-many, ask-for-one} variant, the model retrieves a list of items in an initial turn, after which the user inquires details about one specific element from that list. Finally, in the \textbf{Simple reasoning} variant, the final user follow-up question requires some logical inference or calculation where the previously retrieved information serves as premise. In all four cases, the correctness of the assistant's answer depends on whether the earlier retrieved information continues to hold true given the passage of time.

For the \emph{read+write} setting, we similarly define four variants for diverse coverage. In the \textbf{Repeat after failure} variant, a prior write action failed, and the user later repeats the same (or a similar) request, which, absent any state changes, would again fail. In the \textbf{User confirmation} variant, a prior write action succeeded, and (optionally after intervening turns) the user subsequently asks whether the result still holds (i.e., ``Is~X in state~Y?'', e.g., ``Am I successfully booked on the flight?''). In the \textbf{Repetition of the same request} variant, a write action succeeded earlier, but the user issues the identical request again after intervening turns, as if forgetting the earlier outcome. Finally, in the \textbf{In-context availability / state change} variant, a read action via tool-call exposes a limited resource (e.g., capacity or slots). The user’s successive actions consume that resource until exhaustion, at which point the user issues a request that implicitly exceeds the remaining capacity.

Based on this taxonomy, we manually authored a single exemplar multi-turn trajectory for \emph{each} variant within a scenario (totaling four exemplars per scenario). These served as in-context examples for \texttt{GPT-4o}~\cite{openai2024gpt4technicalreport}, which was prompted to synthetically generate candidate trajectories under the same or similar available set of functions/tools. More details are shown in Appendix~\ref{append:traj_constuct_prompt}.

\subsection{Filtering and quality assurance}
We applied a two-stage filtering pipeline to ensure the quality of the synthetic trajectories. First, we used \texttt{GPT-4.1} as an LLM-as-judge to automatically filter out low-quality outputs according to the following rules: (i) the final user question must not contain explicit instructions for tool invocation, (ii) user questions must not suffer from missing premises, which means the information provided by the previous tool call must be sufficient and necessary for answering the user's question, and (iii) trajectories must exhibit genuine time dependence such that ignoring the elapsed time would lead to misalignment with human expectations regarding tool call decisions. The prompt for the LLM judge is shown in Appendix~\ref{append:llm-judge}.

Second, we conducted detailed human inspection of all retained trajectories. We manually checked for incorrect role orderings, formatting inconsistencies, hallucinated content in earlier turns, and violations of the same criteria applied during the automatic stage. We also ensure that when there are human names, only fictitious or celebrity names are included. After both rounds of filtering, we obtained \textbf{1864 high-quality trajectories}. An example trajectory can be found in Figure~\ref{fig:labeling_view}.

\subsection{Adding time stamps to messages}
\label{subsec:tictoc_add_ts}
To evaluate temporal awareness, every message in a trajectory is assigned a specific timestamp representing the time at which that message was completed. The generation pipeline follows a chronological order: establishing a start time, simulating realistic delays for intermediate turns, and injecting controlled time gaps before the final user query. All timestamps are formatted according to ISO 8601.

\paragraph{Initial timestamp determination}
First, we determine a logical starting date and time for the conversation. We employ \texttt{GPT-4o} to analyze the context of each scenario and generate a timestamp for the first user message that ensures the scenario is temporally consistent (e.g., ensuring that one would only book hotels for future check-in dates). Human inspection on a subset of samples confirmed temporal consistency. 

\paragraph{Intermediate message timestamps}
For all subsequent messages \emph{except} the final user turn, we simulate realistic inter-message delays using a lightweight interaction-time model. We denote a Gaussian distribution with mean $\mu$ and standard deviation $\sigma$, truncated to the interval $[a,b]$, as $T\mathcal{N}(\mu,\sigma,[a,b])$. Similarly, $T\mathcal{LN}$ denotes a truncated Log-Normal distribution.

To simulate the time required for reading, writing, and model generation, we sample three per-trajectory pace variables:
\begin{itemize}
    \item User reading speed $r$ (words/minute) $\sim T\mathcal{N}(\mu_r,\sigma_r,[a_r,b_r])$.
    \item User writing speed $w$ (words/minute) $\sim T\mathcal{LN}(\mu_w,\sigma_w,[a_w,b_w])$.
    \item System generation speed $g$ (words/second) $\sim T\mathcal{N}(\mu_g,\sigma_g,[a_g,b_g])$.
\end{itemize}
Parameters used for these distributions are presented in Appendix~\ref{append:time_param}. Time deltas are computed as follows: user messages induce a writing time proportional to their word count and $w$; assistant messages induce generation time proportional to their word count and $g$; and tool messages induce a \verb|~|1-second execution time. When a user message follows an assistant response, we add reading time proportional to the assistant's previous message length and $r$. A small random jitter $\varepsilon\sim\mathcal{N}(0,\sigma_\varepsilon^2)$ is added to all estimates to prevent artificial regularity. All these time deltas are added sequentially from the initial timestamp to form the timestamps for all intermediate messages.

\paragraph{Final user message timestamp}
To measure sensitivity to varying amounts of time elapses, we construct three distinct versions of each trajectory. These versions share identical time stamps up to the penultimate turn but differ in the elapsed time interval ($\Delta t$) preceding the final user question.

We categorize the magnitude of $\Delta t$ into three levels: \emph{Small}, \emph{Medium}, and \emph{Large}. The actual duration for each level is context-dependent and governed by the scenario's inherent sensitivity (Low, Medium, or High). For instance, a `Large' time gap in a high-frequency trading scenario differs significantly from that in a slow-moving archival scenario.
Specifically, we sample $\Delta t$ from a truncated Gaussian distribution $T\mathcal{N}$ using the parameters corresponding to the scenario's sensitivity and the target time-gap level, as detailed in Table~\ref{tab:params}, Appendix~\ref{append:time_param}. These sampled intervals are then added to the preceding timestamp to produce the final ISO 8601 timestamps.


\vspace{-0.1cm}
\subsection{Preference collection and aggregation}
\label{sec:human-preference-collection}
We treat each trajectory-timestamp pair as an independent sample, yielding a total of $1864 \times 3 = 5592$ samples. Human annotators were asked to assess, for each sample and its associated context, which action was preferable: (i) directly answering without invoking any tool (\emph{Direct}); (ii) calling a tool and answering based on updated information (\emph{Tool}); (iii) uncertainty with a slight preference for direct answering (\emph{Lean-Direct}); or (iv) uncertainty with a slight preference for tool use (\emph{Lean-Tool}). Additional details of the preference collection procedure are provided in Appendix~\ref{append:pref_collect_details}.

To aggregate annotations, we assign numerical scores to each option: 0 for \emph{Direct}, 1 for \emph{Lean-Direct}, 2 for \emph{Lean-Tool}, and 3 for \emph{Tool}. For each sample, we compute the mean score across annotators. Samples with mean scores $S$ between 0.5 and 2.5 indicate substantial annotator uncertainty regarding whether tool use is preferable; such samples are excluded from evaluation. This filtering results in \textbf{3016 retained samples}, comprising 1112 \emph{prefer-Tool} samples with mean scores less than 0.5 (where annotators are confident that tool use is preferable) and 1904 \emph{prefer-noTool} samples with mean scores greater than 2.5 (where annotators are confident that direct answering is preferable). The overall inter-annotator agreement, measured by Krippendorff’s alpha coefficient, is 0.8574, indicating high reliability.



\section{Experiments}
\label{sec:experiments}
In this section, we evaluate how well a variety of contemporary LLM agents align their tool-use decisions with human time perception on \dsname, and make both prompting and post-training alignment efforts to mitigate the misalignment.

\begin{figure*}[h]
    \centering
    \includegraphics[width=1\linewidth]{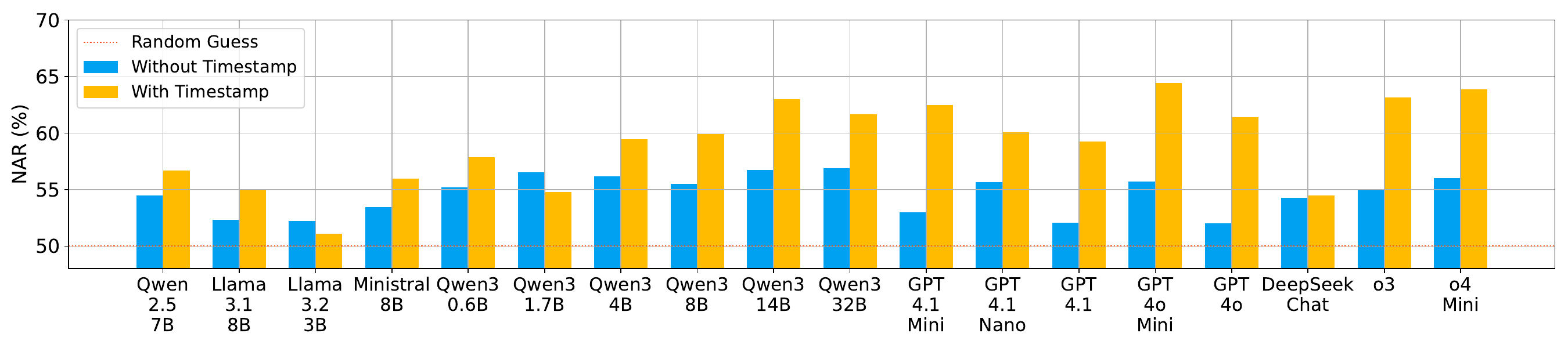}
    \caption{\textbf{Normalized alignment rate of models with and without timestamps.} Without timestamps, models perform only slightly above random (max alignment marginally exceeding 55\%). With timestamps, larger commercial models improve modestly, peaking no more than 65\%.}
    \vspace{-8pt}
    \label{fig:w_wo_timestamp}
\end{figure*}
\begin{figure*}[h]
    \centering
    \includegraphics[width=1\linewidth]{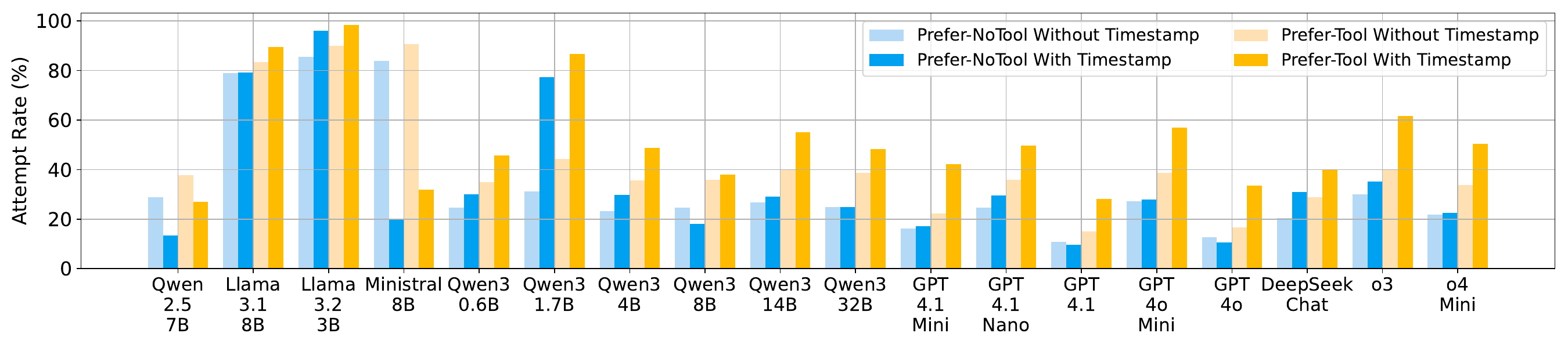}
    \caption{\textbf{Model attempt rates for both \emph{prefer-Tool} and \emph{prefer-noTool} cases.} Without timestamps, models show varying tool-use biases; with timestamps, attempt rates often rise for both classes, indicating limited alignment with human-like temporal awareness.}
    \vspace{-12pt}
    \label{fig:per_class_attempt_rate}
\end{figure*}

\paragraph{Evaluation setup} For open-weight models, we have full control over the chat template. We therefore consider a deployment scenario where an LLM agent is deployed on some system, and timestamps are provided to the model from the system by inserting the system wall-clock time at the beginning of each user, assistant, and tool message. For example, in Qwen models, a user message begins with \texttt{<|im\_start|>user\textbackslash n}. When timestamps are included, this prefix becomes something like \texttt{<|im\_start|>user\textbackslash n[2025-12-04T10:22:44Z]}. 
For proprietary models, we do not have access to their prompt templates or tokenizers. To achieve a comparable effect, we therefore prepend the timestamp string to the beginning of each message’s textual content. In Appendix~\ref{append:explicit_deltaT_grounding}, instead of presenting the absolute timestamp to the LLM, we include a slightly different setting where the explicit value of $\Delta t$ is given. This simplifies the problem by delegating the task of
keeping track of elapsed time to the host system.
All models are evaluated with \texttt{Temperature}=0, where applicable. For Qwen3 models operating in reasoning mode, we follow the officially recommended settings and instead use \texttt{Temperature}=0.6, \texttt{TopP}=0.95, \texttt{TopK}=20, and \texttt{MinP}=0.

\paragraph{Metrics} The main metric we use is 
the \textit{Normalized Alignment Rate} ($NAR$), which is defined as $NAR=\frac{1}{2} \left( \frac{TP}{TP + FN} + \frac{TN}{TN + FP} \right)$. Here, $TP$ refers to \emph{prefer-Tool} samples on which the model attempted a tool call, $TN$ refers to \emph{prefer-noTool} samples on which the model did not attempt a tool call, $FP$ refers to \emph{prefer-noTool} samples on which the model attempted a tool call, and $FN$ refers to \emph{prefer-Tool} samples on which the model did not attempt a tool call. Based on the definition, a 50\% $NAR$ is equivalent to random guessing.
When comparing model behaviors on \emph{prefer-Tool} and \emph{prefer-noTool} samples, we report the \textit{Attempt Rate}\footnote{As our focus is on alignment between models’ tool-call decisions and human time perception, every tool-call attempt (i.e., whenever a model decides to invoke a tool) counts toward the attempt rate. However, not all tool calls are accurate; models may pass incorrect arguments or hallucinate tool names.}, defined as the proportion of samples on which a model attempted a tool call.

\subsection{How well do LLMs' tool-use decisions align with human time perception?}
\label{sec:w_wo_timestamp}

We evaluate 18 proprietary and open-weight LLMs\footnote{Instruction-tuned versions are used for Qwen2.5, Ministral, and Llama models.}, both with and without access to timestamps, and report the Normalized Alignment Rate in Figure~\ref{fig:w_wo_timestamp}. For each sample, the model’s behavior is considered \emph{aligned} if its tool-call decision matches the human preference judgment collected in Section~\ref{sec:human-preference-collection}.

It can be seen that without temporal information, as expected, most models perform similar to random guessing, with the highest normalized alignment rate reaching just above 55\%. When timestamps are provided, proprietary OpenAI models~\cite{openai2024gpt4technicalreport, openai_o3_o4_mini2025} and some larger Qwen3 models~\cite{yang2025qwen3} show a noticeable improvement. However, the overall alignment rate remains low, with no models achieving an $NAR$ of more than 65\%. Even in the simplified setting in Appendix~\ref{append:explicit_deltaT_grounding}, little alignment improvement is observed. To further analyze models' failure modes in slow and fast-changing environments, we leverage our stratification of scenario sensitivity levels and break down the results by this dimension in Figure~\ref{fig:per_sensitivity_level_nar}. It can be seen that models fail uniformly across slow and fast-changing environments. Though models on average fail slightly less on medium-sensitivity scenarios, the difference is very marginal and the overall alignment rate is low.

We report the Attempt Rate separately for the \emph{prefer-Tool} and \emph{prefer-noTool} cases in Figure~\ref{fig:per_class_attempt_rate}. Without timestamps, most models exhibit higher attempt rates on \emph{prefer-Tool} samples than on \emph{prefer-noTool} samples, but each model displays a distinct bias in tool-use tendencies. For example, \texttt{Ministral-8B} and \texttt{Llama-3.2-3B} tend to invoke tools on nearly all samples, whereas OpenAI and Qwen models tend to refrain from invoking tool calls in most cases. 
 With timestamps provided, one would expect human-like temporal awareness to manifest as an increased attempt rate on \emph{prefer-Tool} samples and a decreased attempt rate on \emph{prefer-noTool} samples. However, we observe that attempt rates increase across both subsets for most models. 
 This pattern further indicates that current models struggle to effectively exploit temporal information and fail to align their tool-use decisions with human time perception.  


\vspace{-0.15cm}
\subsection{The correlation with conversation length}
\label{sec:role_len}

\begin{figure}[h]
    \centering
    \includegraphics[width=1\linewidth]{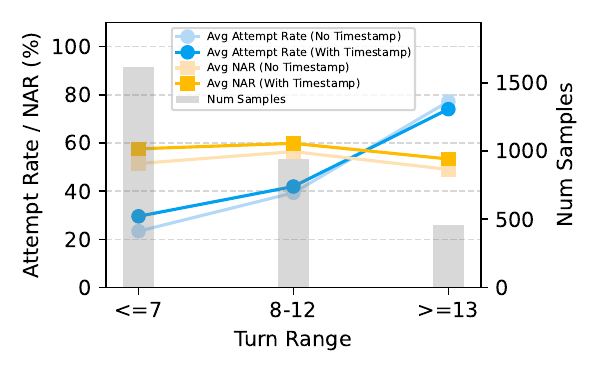}
    \vspace{-0.5cm}
    \caption{\textbf{Attempt rate, normalized alignment rate, and sample distribution across `short', `medium', and `long' trajectories.} Results are averaged across all models. A positive correlation between conversation length and tool-call frequency can be observed, paired with a dip in $NAR$ on longer trajectories.}
    \label{fig:avg_model-turn}
    \vspace{-0.5cm}
\end{figure}

To investigate the impact of conversation length on model performance, we categorized the retained samples into three groups based on trajectory length: short ($\le$ 7 turns), medium (8--12 turns), and long ($\ge$ 13 turns). For each subset, we independently calculated the attempt rate and normalized alignment rate. 
Figure~\ref{fig:avg_model-turn} presents the results averaged across all 18 models, while comprehensive per-model breakdowns are provided in Figure~\ref{fig:permodel_role_of_len}. To avoid potential artifacts arising from correlations between trajectory length and elapsed time, we further design a controlled analysis that disentangles turn count from temporal gaps in Appendix~\ref{append:more_result_turn}.

Our analysis reveals a positive correlation between conversation length and tool-call frequency: models tend to increase their attempt rates as the number of turns grows, regardless of whether timestamp information is provided. As a result, a dip in $NAR$ can be observed for long trajectories. This observation suggests that models may intuitively use ``conversation turns'' as a heuristic for the ``staleness'' of prior observations, rather than effectively leveraging the given explicit time information.

\vspace{-0.15cm}
\subsection{The role of reasoning and why it fails}
\label{sec:long-cot-reason}

\begin{figure}[!h]
    \centering
    \vspace{-8pt}
    \includegraphics[width=\linewidth]{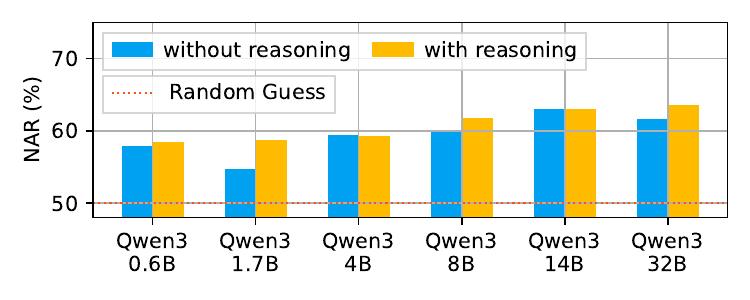}
    \caption{\textbf{Normalized alignment rate of Qwen3 models with and without long CoT reasoning, under settings of both with and without timestamp.} Long CoT reasoning shows no meaningful improvement in tool-use alignment with human time perception.}
    \vspace{-0.4cm}
    \label{fig:qwen3-reason}
\end{figure}

We investigate whether reasoning can enhance models’ ability to leverage temporal information and better align their decisions with human preference. In Figure~\ref{fig:qwen3-reason}, we report the normalized preference alignment rate of Qwen series models in both reasoning and non-reasoning modes.
The results show that reasoning yields only marginal or no improvement in alignment rate. This suggests that while reasoning is powerful for many complex tasks, it does not improve temporal awareness. However, the content of the reasoning traces provides us with an opportunity to understand what the models think and why they would fail.

\paragraph{Absence of temporal information in reasoning} A surprising finding from our analysis of reasoning traces is the frequent omission of timestamps and temporal keywords. Although prior works have demonstrated that models can perform considerably well on explicit temporal reasoning tasks~\cite{chu2024timebenchcomprehensiveevaluationtemporal,yuan2023futureexplainabletemporalreasoning,TempCoT}, in our setting, where temporal dependence is implicit yet critical for decision-making, models often fail to incorporate temporal cues into their rationale. 

\begin{table}[h]
\centering
\small
\resizebox{\linewidth}{!}{
\begin{tabular}{lccc}
\toprule
\textbf{Model} & \textbf{Timestamp} & \textbf{KW 'Timestamp'} & \textbf{KW about Time} \\
\midrule
Qwen3-0.6B-Reason & 31 (1.03\%) & 5 (0.17\%)  & 270 (8.95\%) \\
Qwen3-1.7B-Reason & 58 (1.92\%) & 14 (0.46\%)  & 314 (10.41\%) \\
Qwen3-4B-Reason & 48 (1.59\%) & 21 (0.70\%)  & 358 (11.87\%) \\
Qwen3-8B-Reason & 39 (1.29\%) & 33 (1.09\%)  & 477 (15.82\%) \\
Qwen3-14B-Reason & 75 (2.49\%) & 35 (1.16\%)  & 448 (14.85\%) \\
Qwen3-32B-Reason & 96 (3.18\%) & 43 (1.43\%)  & 382 (12.67\%) \\
\bottomrule
\end{tabular}
}
\caption{\textbf{Occurrences and percentages of timestamp and time-related keywords (KW) found in the reasoning traces of different Qwen3 models.} Timestamp refers to the timestamp of standard format, while KW 'Timestamp' refers to the exact keyword 'timestamp', and KW about Time refers to the keywords about time occurred in the reasoning traces.}
\label{tab:reason_time}
\vspace{-0.5cm}
\end{table}

Table~\ref{tab:reason_time} presents the frequency of timestamps and time-related terms in Qwen3 reasoning traces. Timestamps appear in fewer than 4\% of traces, and explicit mentions of the term ``timestamp'' occur in less than 1.5\%. Even broader temporal keywords (e.g., ``time'', ``date'', ``hour'') appear in under 15\% of cases. These results suggest a significant gap: while models possess temporal reasoning capabilities, they struggle to spontaneously deploy them in practical scenarios, indicating a misalignment with human-like temporal perception.

\paragraph{Think-Answer mismatches in reasoning} Recent studies~\cite{shen2025mitigatingthinkanswermismatchllm} highlight that LLMs occasionally exhibit think-answer mismatches, where a model's internal reasoning contradicts its final output. We observe this phenomenon to be particularly pronounced in our setting, as the final decision, namely whether to invoke a tool, often diverges from the conclusion reached during reasoning. Table~\ref{tab:think-action-conflicts} quantifies these mismatches and their impact on False Negatives (FN) and False Positives (FP) across Qwen3 reasoning models. We categorize these inconsistencies into two types: \textit{Type 1}, where the model decides to call a tool during reasoning but outputs a direct answer; and \textit{Type 2}, where reasoning concludes with a direct answer, yet the final response initiates a tool call.

\begin{table}[h]
\centering
\small
\resizebox{\linewidth}{!}{
\begin{tabular}{lcccc}
\toprule
\textbf{Model} & \textbf{Type-1} & \textbf{\% of FN} & \textbf{Type-2} & \textbf{\% of FP} \\
 & \textbf{Mismatch} & \textbf{Caused} & \textbf{Mismatch} & \textbf{Caused} \\
\midrule
Qwen3-0.6B-Reason & 2.99\% & 4.11\% & 5.71\% & 20.00\% \\
Qwen3-1.7B-Reason & 0.17\% & 0.20\% & 25.13\% & 61.26\% \\
Qwen3-4B-Reason & 0.03\% & 0.00\% & 5.94\% & 19.75\% \\
Qwen3-8B-Reason & 0.37\% & 0.51\% & 1.99\% & 6.09\% \\
Qwen3-14B-Reason & 0.17\% & 0.38\% & 1.79\% & 4.35\% \\
Qwen3-32B-Reason & 0.27\% & 0.57\% & 1.99\% & 7.88\% \\
\bottomrule
\end{tabular}
}
\caption{\textbf{Think-Answer mismatch rates and their impact on False Negatives (FN) and False Positives (FP) across Qwen3 reasoning models.} Type 2 mismatches substantially account for FP errors, whereas Type 1 mismatches are mostly negligible.}
\label{tab:think-action-conflicts}
\vspace{-0.1cm}
\end{table}

\begin{figure*}[h]
    \centering
    \includegraphics[width=1\linewidth]{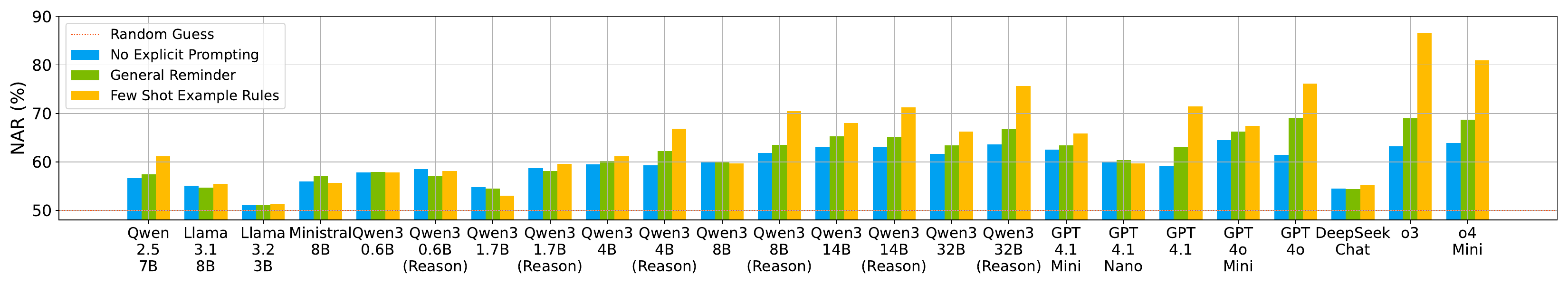}
    \caption{\textbf{Normalized alignment rate of all models with and without prompting-based alignment efforts.} Detailed prompting with few-shot example rules yields notable boost in alignment rate for advanced reasoning models (e.g. \texttt{o3}, \texttt{o4-mini}), but has limited effect on most others.}
    \vspace{-0.2cm}
    \label{fig:all_model_ar_with_prompt}
\end{figure*}
Our analysis reveals that Type 2 mismatches significantly hinder alignment with human preference. Notably, for the 1.7B model, Type 2 mismatches account for 61\% of False Positive (FP) errors. The 0.6B and 4B models also exhibit substantial impact, with approximately 20\% of FPs stemming from this inconsistency. Model scaling appears to reduce the prominence of this issue: for larger models, the mismatch rate falls below 2\%, and its contribution to FP errors decreases to under 8\%. In contrast, Type 1 mismatches are negligible except for the 0.6B model, indicating that models rarely fail to execute a tool call once the reasoning process has committed to it.

\vspace{-0.15cm}
\subsection{Alignment efforts}
\label{sec:mitigation}

Given the substantial divergence between the tool-use decisions of models and human preferences, we explored both prompting strategies and post-training to improve alignment.

\vspace{-0.1cm}
\paragraph{Prompting strategies} We first tested a minimal intervention by adding a general reminder to the system prompt: \textit{``Note that the environment may be dynamic. Be aware of the time elapsed.''} This intervention had little to no effect. We then designed a stronger prompt that included few-shot examples illustrating rules for when tool calls are preferable or unnecessary depending on the amount of elapsed time. The complete instruction is shown in Figure~\ref{fig:rule}. All results are reported in Figure~\ref{fig:all_model_ar_with_prompt}. 
The results reveal that for advanced reasoning models such as \texttt{o3} and \texttt{o4-mini}, detailed prompting with example rules yields a substantial boost in alignment rate. However, for most other models, prompting-based strategies show marginal or no effectiveness. This finding suggests that, similar to prior alignment efforts in reducing harmful outputs and mitigating jailbreaks~\cite{yi2024jailbreakattacksdefenseslarge}, effective alignment of tool-use decisions in temporally dynamic environments requires targeted post-training rather than prompt engineering alone.  

\begin{figure}[h]
    \centering
    \includegraphics[width=1\linewidth]{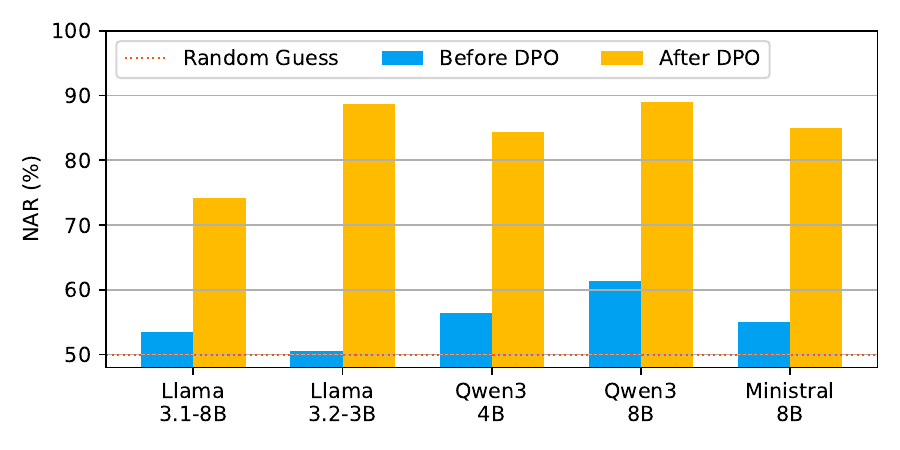}
    \caption{\textbf{Normalized alignment rate of selected open-source models before and after DPO tuning.} Massive alignment gains are achieved across all models.}
    \label{fig:dpo_effect}
    \vspace{-0.6cm}
\end{figure}

\vspace{-0.1cm}
\paragraph{Post-training with DPO} We split \dsname~by scenario into training and testing sets, resulting in an approximate 65\%:35\% split in terms of number of samples (see Appendix~\ref{append:dpo_data_split} for the detailed scenario split). Conceptually similar to ODPO~\cite{amini2024directpreferenceoptimizationoffset}, we employed Direct Preference Optimization (DPO) with a dynamic margin to fine-tune selected open-source models on the training split for a single epoch. Detailed configurations are provided in Appendix~\ref{append:DPO_config}.
The results, reported in Figure~\ref{fig:dpo_effect}, demonstrate massive alignment gains across all trained models. This confirms the strong potential of targeted post-training as a necessary step for achieving reliable temporal alignment in multi-turn tool-use models.

\vspace{-0.2cm}

\section{Conclusion}
We identify \textit{temporal blindness} as a critical limitation of multi-turn LLM agents: models often fail to account for the passage of real-world time between messages when making tool-call decisions, leading to either over-reliance or under-reliance on prior context. To evaluate this, we introduced \dsname, a diverse dataset of multi-turn user–agent conversation trajectories that include tool calls. By evaluating 18 open-weight and proprietary models and analyzing their failure modes, we underscore the misalignment between agents' tool-call decision with human time perception. Our data and findings provide a first step toward understanding and mitigating temporal blindness, offering insights to foster the development of more time-aware and human-aligned agents.

\section*{Limitations}

The scenarios and conversation trajectories in \dsname~focus on tool use and user–agent interactions in the text-only setting. Extending the dataset to multimodal tool-use scenarios (e.g., image retrieval or vision–language tools) is a natural direction for future work. In addition, our DPO experiments are limited to open-source models with at most 8B parameters due to computational constraints. Applying targeted DPO to larger-scale models could provide further insights into the effectiveness of post-training for aligning agent tool-use decisions with human temporal perception.

\section*{Acknowledgments}

This work was supported in part by NSF CAREER Award 1942230, the ONR PECASE Award N00014-25-1-2378, ARO Early Career Program Award 310902-00001, Army Grant W911NF-21-2-0076, NSF Award CCF-2212458, NSF Award 2229885 (NSF Institute for Trustworthy AI in Law and Society, TRAILS), MURI Grant 14262683, DARPA AIQ Grant HR00112590066, and a Meta Research Award 314593-00001.

\bibliography{acl_latex}

\clearpage
\appendix

\section{More details on dataset curation}
\subsection{Taxonomy of scenarios}
\label{append:taxonomy}
A complete inventory of the scenarios in \dsname~is provided in Table~\ref{tab:scenario_invetory}, which details the sensitivity level and trajectory count for each scenario.

\subsection{Synthetic trajectory construction}
\label{append:traj_constuct_prompt}
After manually authoring an exemplar trajectory for each variant in every scenario, we prompt \texttt{GPT-4o} using the instructions shown in Figure~\ref{fig:prompt_r} and Figure~\ref{fig:prompt_rw} for the \emph{read-only} and \emph{read+write} settings, respectively. We synthetically generate 50 candidate trajectories per scenario, allowing ample room for quality filtering in the next stage. To achieve this, we uniformly sample a ``preferred\_strategy'' (corresponding to one specific variant) for each generation instance, aiming at achieving a balanced distribution among the 4 variants. However, the prompt context includes the definitions of all four variants. This design grants \texttt{GPT-4o} the flexibility to determine if a specific variant is unsuitable for the current scenario, allowing it to pivot to a more appropriate strategy to avoid generating implausible or unrealistic interactions.

\subsection{Prompt for \texttt{GPT-4.1} during first-stage automatic quality filtering}
\label{append:llm-judge}
Before conducting detailed human inspection on each candidate trajectory, we configure \texttt{GPT-4.1} as an LLM judge to conduct a first-stage automatic quality filtering. The detailed instruction and criteria are shown in Figure~\ref{fig:verify}.

\subsection{$T\mathcal{N}$ and $T\mathcal{LN}$ parameters used for timestamp addition}
\label{append:time_param}

To simulate realistic inter-message delays in user-agent conversations, we model the time elapsed between consecutive messages based on simple assumptions about user and system speeds. Specifically, for each conversation trajectory, we assume a user with a reading speed $r$ and a writing speed $w$, and a model with a text generation speed $g$.

Prior work has shown that human typing speeds are better characterized by a right-skewed distribution rather than a symmetric Gaussian~\citep{10.1145/3173574.3174220}. To account for this asymmetry, we model user writing speed using a truncated log-normal distribution. Both user reading speed and model generation speed are approximated by truncated normal distributions.

Formally, we sample the three speed variables as follows:
\begin{itemize}
    \item User reading speed $r$ (words per minute) $\sim T\mathcal{N}(\mu_r, \sigma_r, [a_r, b_r])$.
    \item User writing speed $w$ (words per minute) $\sim T\mathcal{LN}(\mu_w, \sigma_w, [a_w, b_w])$.
    \item System generation speed $g$ (words per second) $\sim T\mathcal{N}(\mu_g, \sigma_g, [a_g, b_g])$.
\end{itemize}

The means, standard deviations, and lower and upper truncation bounds for these distributions are reported in Table~\ref{tab:timestamp-pace-params}. Given a sampled speed, the time required for reading, writing, or generation is computed as a linear function of the corresponding text length.


\begin{table}[h]
\centering
\small

\begin{tabular}{lcccc}
\toprule
\textbf{Rate} & $\mu$ & $\sigma$ & $a$ & $b$ \\
\midrule
Read ($r$, wpm) & 238 & 60 & 50 & $\infty$ \\
Write ($w$, wpm) & 3.61 & 0.40 & 5 & $\infty$ \\
Generate ($g$, wps) & 40 & 16 & 10 & $\infty$ \\
\bottomrule

\end{tabular}
\caption{Pace distributions used for non-final message timing. Reading and generation speeds use truncated Gaussians $T\mathcal{N}(\mu,\sigma,[a,b])$, while writing speed uses a truncated log-normal $T\mathcal{LN}(\mu,\sigma,[a,b])$.}
\label{tab:timestamp-pace-params}
\end{table}
\begin{table*}[h]
\centering
\footnotesize
\begin{tabular}{@{}l l l l l l@{}}

\toprule
Sensitivity & Elapse & {Mean ($\mu$)} & {Std. dev. ($\sigma$)} & {Minimum ($a$)} & {Maximum ($b$)} \\
\midrule
Low & Small & \SI{3}{minutes} & \SI{1}{minute} & \SI{1}{minute} & \SI{6}{minutes} \\
& Medium & \SI{3}{days} & \SI{1}{day} & \SI{1}{day} & \SI{6}{days} \\
& Large & \SI{3}{months} & \SI{1}{month} & \SI{1}{month} & \SI{6}{months} \\
\midrule
Medium & Small & \SI{3}{minutes} & \SI{1}{minute} & \SI{1}{minute} & \SI{6}{minutes} \\
& Medium & \SI{3}{hours} & \SI{1}{hour} & \SI{1}{hour} & \SI{6}{hours} \\
& Large & \SI{3}{days} & \SI{1}{day} & \SI{1}{day} & \SI{6}{days} \\
\midrule
High & Small & \SI{3}{seconds} & \SI{1}{second} & \SI{1}{second} & \SI{6}{seconds} \\
& Medium & \SI{3}{minutes} & \SI{1}{minute} & \SI{1}{minute} & \SI{6}{minutes} \\
& Large & \SI{3}{hours} & \SI{1}{hour} & \SI{1}{hour} & \SI{6}{hours} \\
\bottomrule
\end{tabular}
\vspace{2pt}
\caption{Truncated Gaussian parameters for elapsed time sampling.}
\label{tab:params}
\end{table*}
\begin{figure*}
    \centering
    \includegraphics[width=1\linewidth]{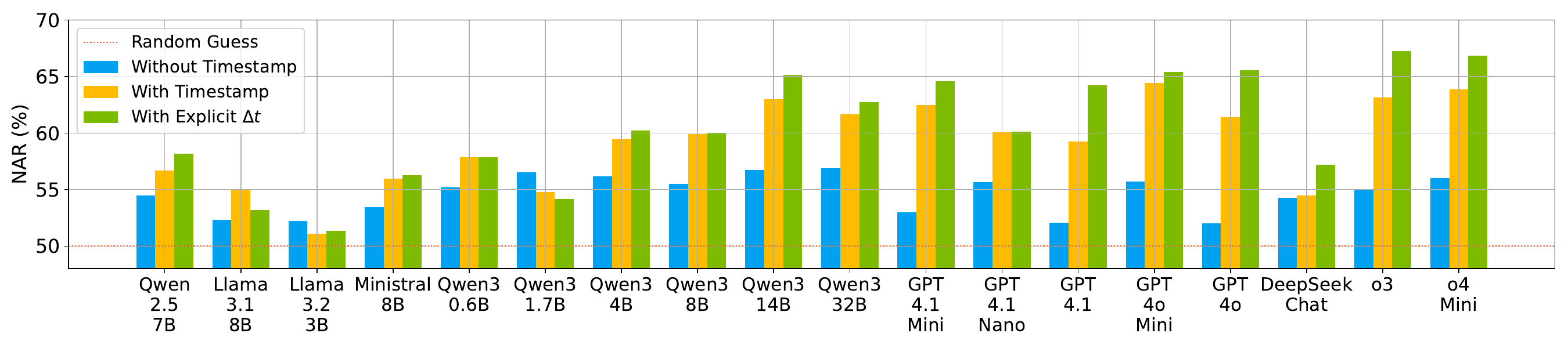}
    \caption{\textbf{Normalized alignment rate of models without timestamps, with ISO 8601 absolute timestamps, and with explicit $\Delta t$ values.} For most models, even with explicit $\Delta t$, there is no alignment increase compared to having only ISO 8601 absolute timestamps. For a few models where there is a little improvement, the improvement is marginal (no more than 1\%-4\%).}
    \label{fig:nar_with_explicit_deltaT}
\end{figure*}
\subsection{Human preference collection}
\label{append:pref_collect_details}

We collected human preferences through a structured survey designed to evaluate agent behavior under different temporal conditions. Each survey consisted of 20 conversation trajectories. For each trajectory, annotators were shown the full conversation along with the three sets of timestamps and were asked to judge which action was more appropriate for the agent at the final turn. The available choices were: (i) directly answering without invoking any tool (\emph{Direct}); (ii) calling a tool and answering based on updated information (\emph{Tool}); (iii) expressing uncertainty with a slight preference for direct answering (\emph{Lean-Direct}); or (iv) expressing uncertainty with a slight preference for tool use (\emph{Lean-Tool}).

\begin{figure*}
    \centering
    \includegraphics[width=0.9\linewidth]{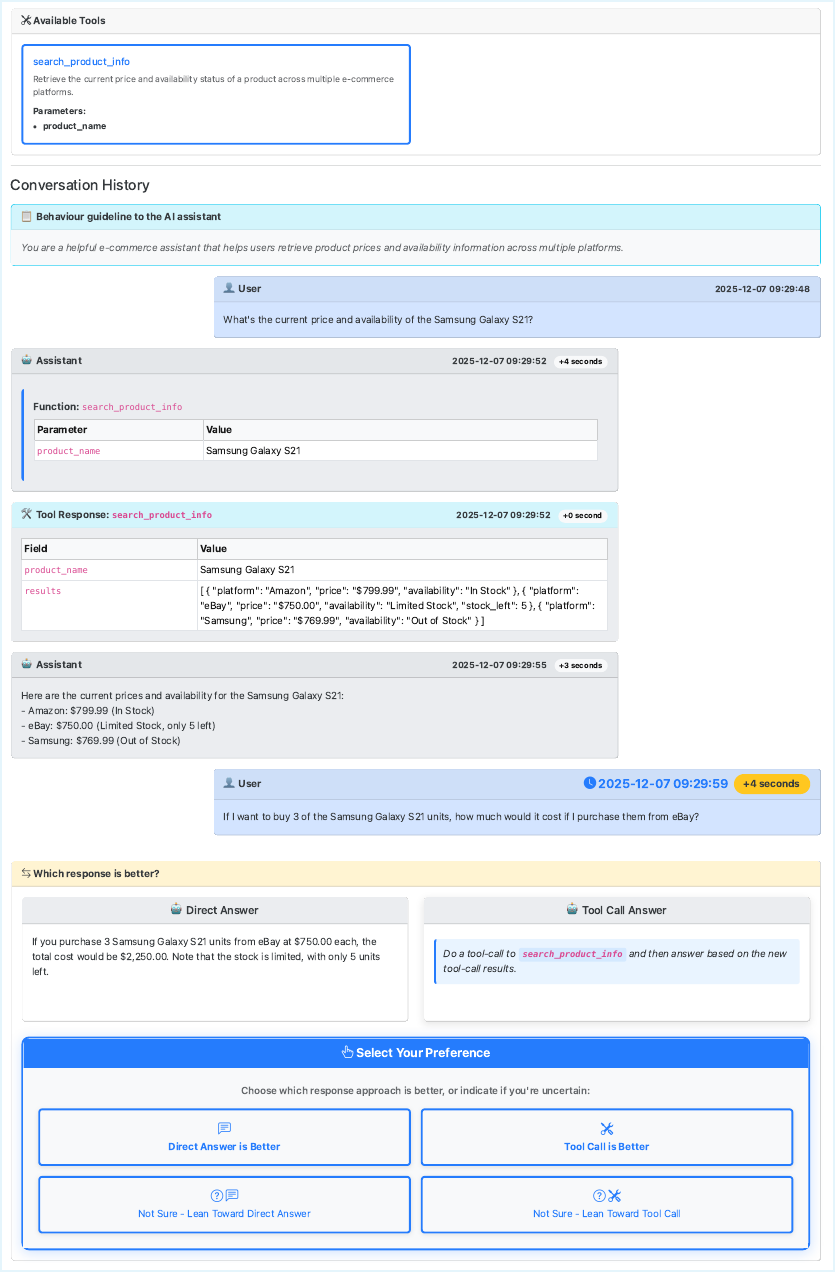}
    \caption{\textbf{Human preference annotation interface.} The interface displays the full conversation history with explicit timestamps, requiring annotators to decide whether the agent should provide a direct answer or perform a tool call based on the temporal context. Preferences are recorded on a four-point scale to capture both definitive choices and uncertainty.}
    \vspace{-3pt}
    \label{fig:labeling_view}
\end{figure*}

\begin{figure}[htbp]
  \centering
    \includegraphics[width=0.8\linewidth]{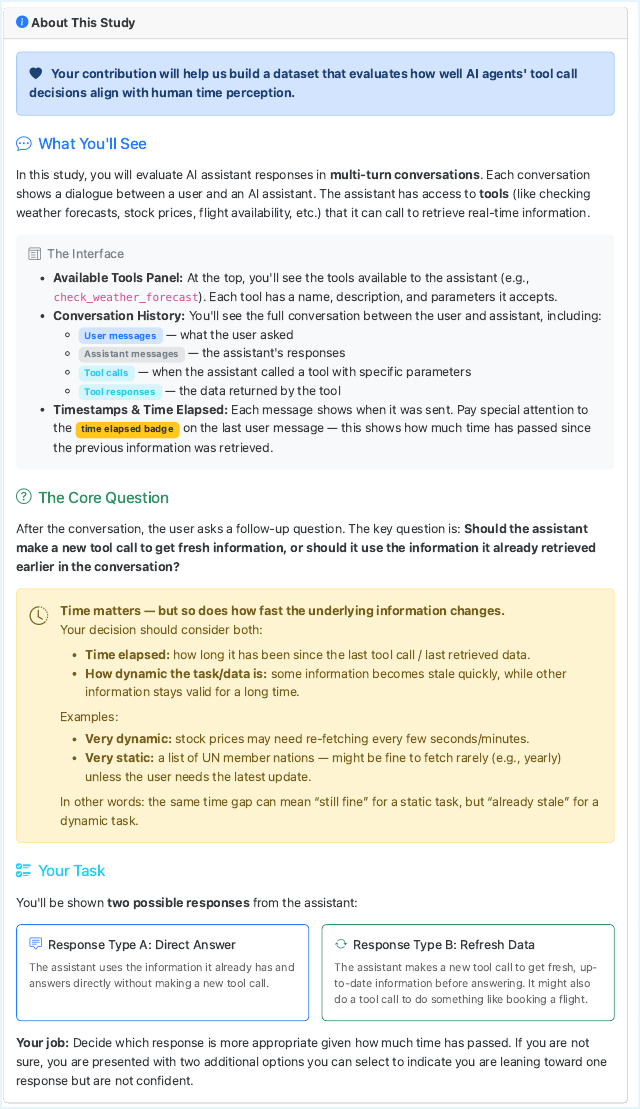}
  \caption{\textbf{The instruction given to the annotators.} It details the interface layout, explains the purpose of data collection and core question of interest, and describes the task requirements.}
  \label{fig:survey-grid}
\end{figure}

Each time an annotator opened the survey, a random set of 20 trajectories was sampled. Annotators were required to provide a username upon submission. This served two purposes: (1) tracking individual contributions for compensation, and (2) ensuring that annotators who completed the survey multiple times were not shown the same trajectories again. This procedure ensured that each trajectory received multiple preference annotations from distinct individuals. A screenshot of the survey interface is shown in Figure~\ref{fig:labeling_view}. The screenshot of the survey instruction page is shown in Figure~\ref{fig:survey-grid}.

Our annotator pool consisted of a mix of undergraduate students, graduate students, and crowd-sourced workers recruited via Credamo\footnote{www.credamo.com}. To ensure annotation quality, we applied a simple but effective sanity check to filter out submissions from annotators who appeared to answer randomly or without understanding the task.
Specifically, for each conversation trajectory, annotators provided preferences under three different time gaps preceding the final user query. While it is expected and acceptable for different annotators to hold different preferences for the same trajectory, the preferences provided by a single annotator should be temporally consistent. As the time gap increases, an annotator’s preference should either remain unchanged or gradually shift toward the \emph{Tool} option, reflecting the increased likelihood that information becomes outdated over time. A preference pattern in which an annotator selects \emph{Tool} for a short time gap but switches to \emph{Direct} for a much longer time gap is considered logically inconsistent.

For each survey submission, we allowed at most one such inconsistency across the 20 trajectories, accounting for the possibility of an occasional typo or mistake. If an annotator exhibited more than two inconsistent cases within a single survey, the entire submission was discarded and no compensation was provided.

Regarding compensation, student annotators received 18 CNY (approximately \$2.6 USD) for each accepted survey. For crowd-sourced workers from Credamo, we accepted the quote of 28 CNY (approximately \$4 USD) per accepted survey completion. The total cost of collecting at least five preference annotations for each sample in \dsname~was approximately \$1,800 USD.

\section{Results with explicit $\Delta t$ values}
\label{append:explicit_deltaT_grounding}
In our main experiments (Section~\ref{sec:experiments}), the temporal information is provided to the model using absolute ISO 8601 timestamps. Here, we simplify the problem by directly presenting the explicit $\Delta t$ values to the LLM context. For example, in Qwen models, the user message prefix now becomes something like \texttt{<|im\_start|>user\textbackslash n[2025-12-04T10:22:44Z; 2 minutes passed]}. This alleviates the need for the model to reason about the time elapsed between messages, and delegates the task of keeping track of elapsed time to the host system. Expanding our results from Section~\ref{sec:w_wo_timestamp}, we report the normalized alignment rates in Figure~\ref{fig:nar_with_explicit_deltaT}. It can be seen that for most models, even with explicit $\Delta t$, there is no alignment increase. For a few models where there is a little improvement, the improvement is marginal (no more than 1\%-4\%).



\section{More results on the correlation with conversation length}
\label{append:more_result_turn}
\begin{figure}[h]
    \centering
    \includegraphics[width=1\linewidth]{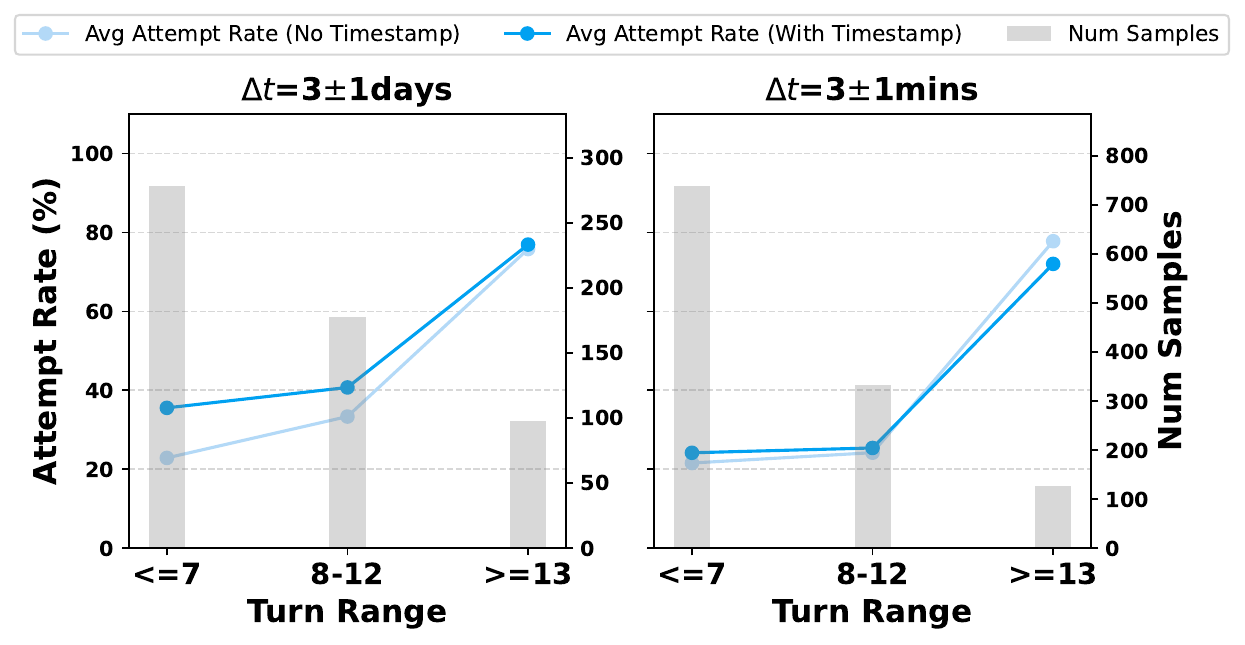}
    \vspace{-0.5cm}
    \caption{\textbf{Attempt rate and sample distribution across `short', `medium', and `long' trajectories with controlled `Long' and `Short' durations.} Results are averaged across all models. A positive correlation between conversation length and tool-call frequency can be observed in both cases.}
    \label{fig:avg_model-turn-controlled}
\end{figure}

To further investigate the impact of conversation length on model performance while avoiding potential artifacts arising from correlations between trajectory length and elapsed time, we design a controlled analysis that disentangles turn count from temporal gaps. Because timestamps are injected with stochasticity (Section~\ref{subsec:tictoc_add_ts}; Appendix~\ref{append:time_param}), obtaining samples with identical elapsed times is difficult. However, since elapsed times follow well-defined distributions (Table~\ref{tab:params}), we grouped samples by the same order of magnitude of elapsed time. We selected: 
 \begin{itemize}
     \item \textbf{``Long Duration''}: Mean elapsed time of 3 days (std 1 day). This corresponds to "medium" elapse in low-sensitivity scenarios and "large" elapse in high-sensitivity scenarios.
     \item \textbf{``Short Duration''}: Mean elapsed time of 3 minutes (std 1 min). This corresponds to "small" elapse in low/medium-sensitivity scenarios and "medium" elapse in high-sensitivity scenarios.
 \end{itemize}
 Within each set, we categorized the samples into three groups based on trajectory length: short ($\le$ 7 turns), medium (8--12 turns), and long ($\ge$ 13 turns). For each group, we independently calculated the attempt rates. 
 Figure~\ref{fig:avg_model-turn-controlled} presents the results averaged across all 18 models.

\begin{figure*}
    \centering
    \includegraphics[width=1\linewidth]{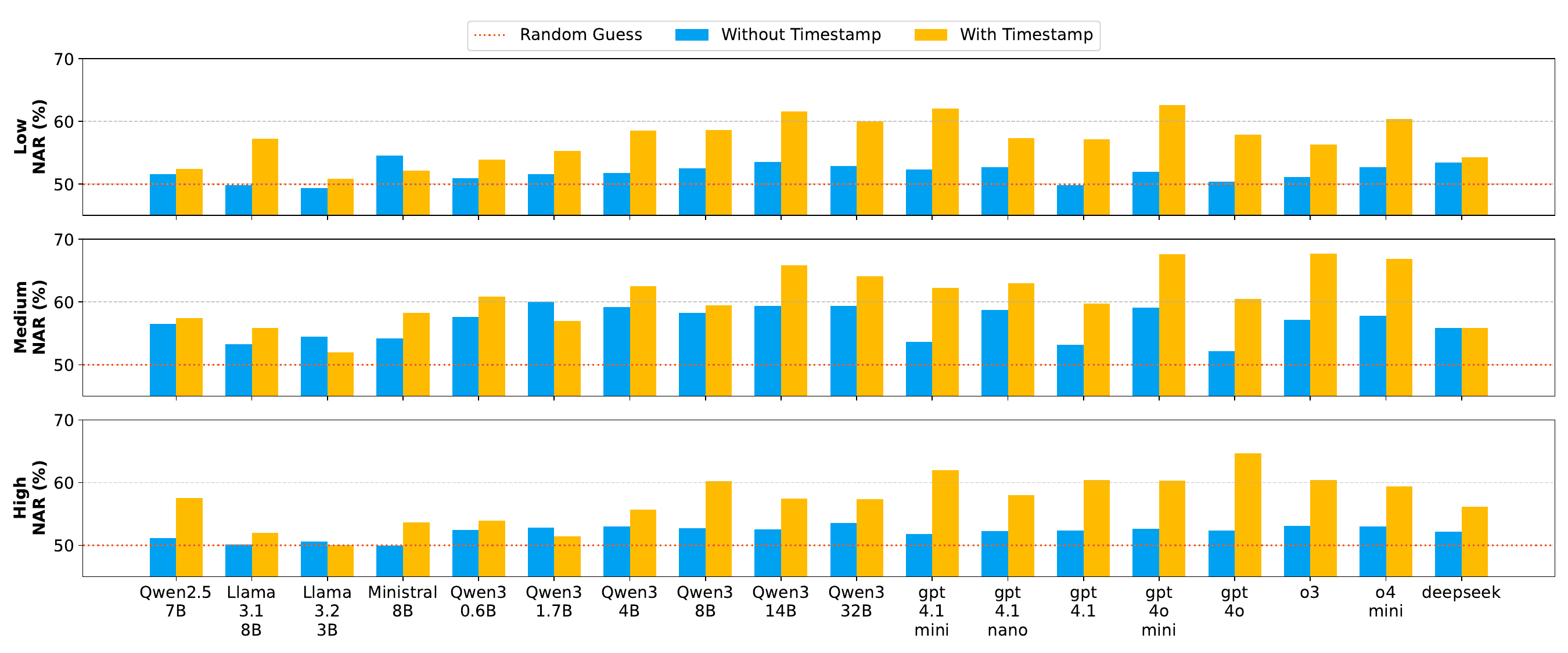}
    \caption{\textbf{Normalized alignment rates for models with and without timestamps across scenarios with low, medium, and high time sensitivity levels.} Scenario stratification follows the categories defined in Section~\ref{subsec:scenario_design}. Overall, models exhibit consistently low alignment rates in both slow- and fast-changing settings. While performance is marginally higher in medium-sensitivity scenarios, the improvement is slight, and overall alignment rate remains poor.}
    \label{fig:per_sensitivity_level_nar}
\end{figure*}

\begin{figure*}
    \centering
    \includegraphics[width=1\linewidth]{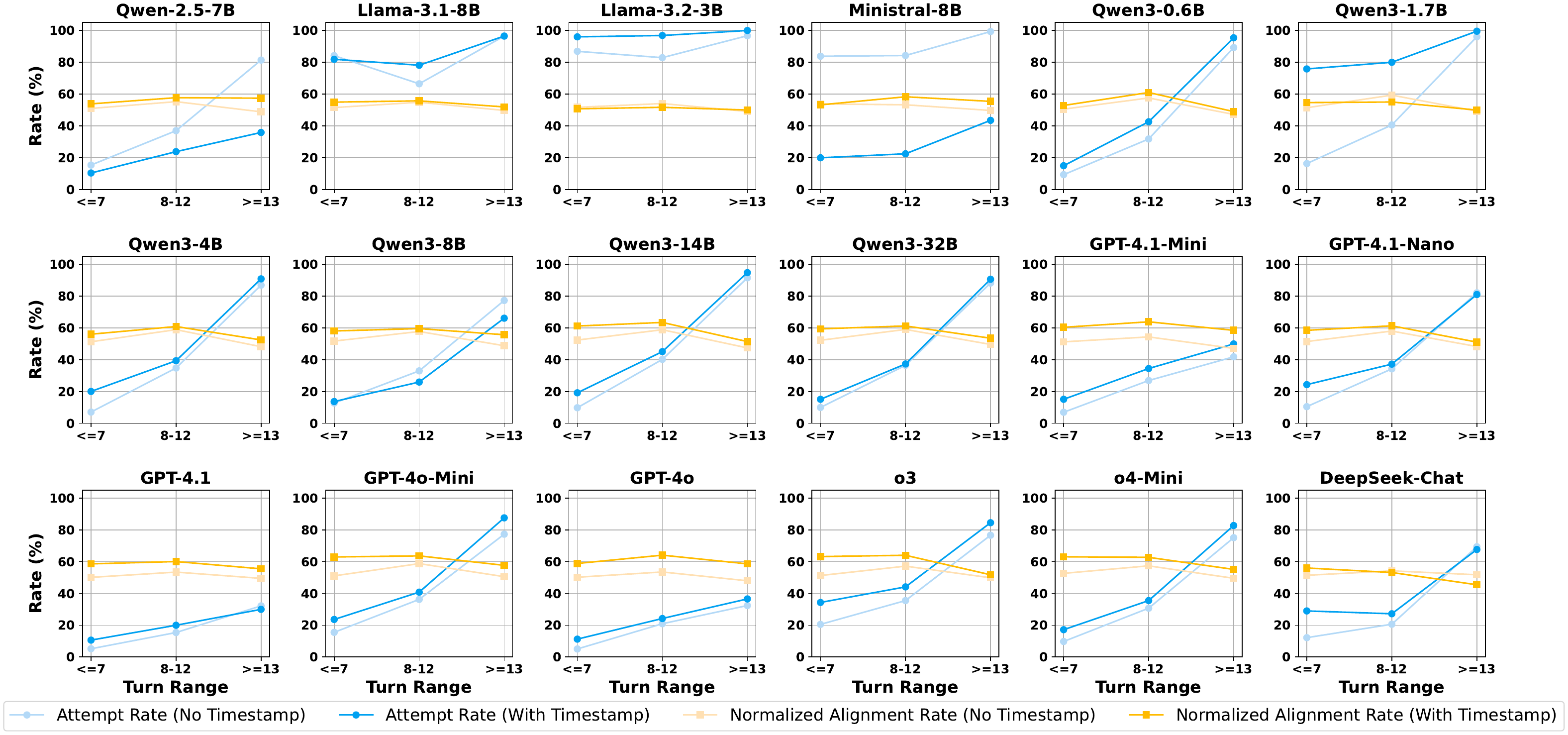}
    \caption{\textbf{Per model attempt rate, normalized alignment rate, and sample distribution across `short', `medium', and `long' trajectories.} A positive correlation between conversation length and tool-call frequency can consistently be observed on all models, paired with a dip in $NAR$ on longer trajectories.}
    
    \label{fig:permodel_role_of_len}
    \vspace{+0.5cm}
\end{figure*}
\begin{table*}[ht]
\centering

\resizebox{\textwidth}{!}{
\begin{tabular}{lccccc}
\toprule
\textbf{Models} & \textbf{Llama-3.1-8B-Instruct} & \textbf{Llama-3.2-3B-Instruct} & \textbf{Ministral-8B-Instruct} & \textbf{Qwen3-4B} & \textbf{Qwen3-8B} \\
\midrule
\textbf{Compute} & \multicolumn{5}{c}{4 $\times$ L40s (distributed training)} \\
\textbf{Precision} & \multicolumn{5}{c}{BF16} \\
\textbf{Optimizer} & \multicolumn{5}{c}{AdamW~\cite{loshchilov2019decoupledweightdecayregularization}} \\
\textbf{Learning Rate} & 5e-7 & 5e-6 & 5e-7 & 5e-6 & 5e-6 \\
\textbf{Beta (in DPO)} & 0.1 & 0.05 & 0.1 & 0.05 & 0.05 \\
\bottomrule
\end{tabular}
}

\caption{Training configurations for different models}
\label{tab:training_configs}
\end{table*}

\section{More details on DPO training}
\subsection{Dataset split}
\label{append:dpo_data_split}

To make post-training interventions to improve alignment, we split \dsname~by scenarios into a training split and a test split. The scenarios that fall within the test split are:

\begin{itemize}
    \item \textbf{Low Sensitivity:} \textit{Airline Baggage Policy}, \textit{Astronomical Object Info}, \textit{Chemical Safety Data Sheet (SDS)}, \textit{City Population Data}, \textit{Office Finder}, \textit{Patent Metadata Lookup}, \textit{Public Health Screening Guidelines}, and \textit{Streaming Service Library Checker}.
    \item \textbf{Medium Sensitivity:} \textit{Airport Security Line Wait Time}, \textit{Hiking Trail Condition Report}, \textit{Hotel Room Availability \& Rates}, \textit{Job Board Live Listings}, \textit{Movie Showtimes Finder}, and \textit{Tide Height Tracker}.
    \item \textbf{High Sensitivity:} \textit{Cloud Server Load Balancer}, \textit{Emergency Alert / 911 Dispatch Interface}, \textit{ICU Vitals Monitor}, \textit{Live Auction Price Retrieval}, \textit{Live Sports Score Tracker}, \textit{Live Vehicle GPS Tracker}, \textit{Live Weather Sensor Data}, \textit{Restaurant Wait Time Checker}, \textit{Ride-Hailing Dispatch}, \textit{Stock Market Order Book}, \textit{Train Delay Tracker}, and \textit{Urban Parking-Spot Reservation}.
\end{itemize}

The remaining scenarios (as listed in Table~\ref{tab:scenario_invetory}) fall within the train split. This split results in 50 training scenarios (21 low, 19 medium, and 10 high sensitivity) and 26 test scenarios (8 low, 6 medium, and 12 high sensitivity), and a train-test split of approximately 65\%:35\% in terms of number of samples. As we split the data based on scenarios, this ensures the alignment gain obtained after training is actual generalization, rather than just memorization on similar samples.

\subsection{Training Configurations}
\label{append:DPO_config}

We employ Direct Preference Optimization (DPO) with a dynamic margin, conceptually similar to the ODPO framework proposed by~\citet{amini2024directpreferenceoptimizationoffset}. In standard DPO training, the objective is to maximize the log-likelihood of the preferred response $y_w$ relative to the rejected response $y_l$. The loss function is defined as:

\begin{equation*}
\begin{split}
    \mathcal{L}_{\text{DPO}} = -\mathbb{E}_{\mathcal{D}} \Big[ \log \sigma \Big( & \beta \log \frac{\pi_\theta(y_w|x)}{\pi_{\text{ref}}(y_w|x)} \\
    & - \beta \log \frac{\pi_\theta(y_l|x)}{\pi_{\text{ref}}(y_l|x)} \Big) \Big]
\end{split}
\label{eq:dpo_standard}
\end{equation*}

\noindent where $\sigma$ is the sigmoid function and $\beta$ is a hyperparameter controlling the deviation from the reference model $\pi_{\text{ref}}$. 

To account for varying confidence levels in our preference labels, we introduce a margin term $\delta$ into the objective. The modified loss becomes:
\begin{equation*}
\begin{split}
    \mathcal{L}_{\text{MDPO}} = -\mathbb{E}_{\mathcal{D}} \Big[ \log \sigma \Big( & \beta \log \frac{\pi_\theta(y_w|x)}{\pi_{\text{ref}}(y_w|x)} \\ 
    & - \beta \log \frac{\pi_\theta(y_l|x)}{\pi_{\text{ref}}(y_l|x)} - \delta \Big) \Big]
\end{split}
\label{eq:dpo_margin}
\end{equation*}

\noindent This offset $\delta$ effectively shifts the decision boundary of the sigmoid function. By subtracting $\delta$ from the log-probability ratios, we impose a stricter constraint on the policy: it is insufficient for the model simply to prefer $y_w$ over $y_l$; the implicit reward gap must exceed the threshold $\delta$ to minimize the loss. If $\delta = 0$ (indicating a tie), the model is not forced to separate the responses, preventing overfitting on ambiguous pairs. Conversely, a large $\delta$ forces a significant separation for high-confidence pairs.

In our specific setup, we define $\delta$ dynamically based on the mean score $S \in [0, 3]$ described in Section~\ref{sec:human-preference-collection}, where $S=0$ denotes a strong preference for the ``Direct'' response, $S=3$ denotes a strong preference for the ``Tool Call'', and $S=1.5$ represents a tie. We calculate the margin as:
\begin{equation*}
    \delta = |S - 1.5|
\end{equation*}
Consequently, high-confidence samples ($S \in \{0, 3\}$) are assigned a maximum margin of $\delta=1.5$, ensuring the model learns decisive boundaries for clear cases, while ambiguous cases ($S \approx 1.5$) yield a margin near zero. In this way, samples where the score is 1.6 or 1.4 (near ties) will result in a weight near 0.1, meaning they contribute very little to the gradient update. Samples with a score of 0 or 3 will have a weight of 1.5, forcing the model to prioritize getting those "obvious" cases right.

All models are trained on 4 $\times$ L40s GPUs with fsdp parameter offloading. The training configurations for each model are detailed in Table~\ref{tab:training_configs}.

\begin{figure*}[!h]
\begin{tcolorbox}[
  enhanced, 
  colframe=teal!75!black, 
  colback=white, 
  coltitle=white, 
  colbacktitle=teal!75!black, 
  width=\linewidth, 
  arc=2mm, 
  auto outer arc, 
  boxrule=0.5pt, 
  left=10pt, 
  right=10pt, 
  drop shadow={black!50!white},
  top=10pt, 
  bottom=10pt, 
  title=\textbf{Prompt Template for Read-Only Sample Generation}, 
  fonttitle=\bfseries, 
  title code={\node[rounded corners, fill=blue!75!black, draw=none, text=white] at (frame.title) {\textbf{xxx}};}, 
  attach boxed title to top center={yshift=-2mm}, 
  boxed title style={sharp corners, size=small}, 
]
You are a helpful assistant tasked with creating a sample multi-turn chat trajectory for evaluating an agent's temporal awareness. The goal is to determine whether the agent can correctly interpret the passage of time between conversation turns and decide when to reuse previous tool call results versus when to make a new tool call.
\\

You will be provided with a topic and its description. Your task is to generate a chat history between a user and an assistant, where:

- The user makes an initial request that requires a tool call.

- The assistant responds by calling the tool and providing the result.

- After a time gap (implied or explicit), the user makes a follow-up request that could \textbf{necessarily use the same kind of information as before but the result from the tool call would be different if made at a different time from the previous tool call}. Avoid asking exactly the same question as before.
\\

To boost the quality and diversity of the samples, please use a variety of strategies for the follow-up question. Here are some examples:

1. \textbf{Repeated Ask:} Repeats a request for information that was already retrieved in a prior turn.

2.  \textbf{Comparison:} Retrieve information for item A, then for item B. The follow-up question asks for a comparison between A and B. We can also retrieve info for A and compare it with a given number B, in direct or indirect ways.

3.  \textbf{Retrieve Many, Ask for One:} The initial tool call retrieves a list of items. The follow-up question asks for a specific detail or condition about one of them.

4.  \textbf{Mathematical Reasoning:} The retrieved information is a critical premise for a follow-up question that requires some calculation or logical reasoning.\\

Please follow the structure and format of the provided example, including roles, tool calls, and JSON formatting. The last message should be a user message. Do not include any time indicators such as 'still' in the last message.\\

Here is the example:

Topic: [example\_topic]

Description: [example\_description]

Example: [example]\\

Now, create a similar chat trajectory for the following topic.

Topic: [topic]

Description: [description]\\

You can use the following strategy for the follow-up question if it is appropriate: [preferred\_strategy]. Feel free to use other strategies if they are more appropriate.

Output in the same JSON format as above.

\end{tcolorbox}
\caption{\textbf{The prompt used for generating \emph{read-only} samples with an in-context example.} To fully leverage the instruction following capability of the model, we generate the samples for one variant at a time by passing in a \emph{preferred\_strategy}.}
\label{fig:prompt_r}
\end{figure*}
\begin{figure*}[!h]
\begin{tcolorbox}[
  enhanced, 
  colframe=cyan!75!black, 
  colback=white, 
  coltitle=white, 
  colbacktitle=cyan!75!black, 
  width=\linewidth, 
  arc=2mm, 
  auto outer arc, 
  boxrule=0.5pt, 
  left=10pt, 
  right=10pt, 
  drop shadow={black!50!white},
  top=10pt, 
  bottom=10pt, 
  title=\textbf{Prompt Template for Read+Write Sample Generation}, 
  fonttitle=\bfseries, 
  title code={\node[rounded corners, fill=blue!75!black, draw=none, text=white] at (frame.title) {\textbf{xxx}};}, 
  attach boxed title to top center={yshift=-2mm}, 
  boxed title style={sharp corners, size=small}, 
]
You are a helpful assistant tasked with creating a sample multi-turn chat trajectory for evaluating an agent's temporal awareness. The goal is to determine whether the agent can correctly interpret the passage of time between conversation turns and decide when to reuse previous tool call results versus when to make a new tool call.
\\

You will be provided with a topic and its description. Your task is to generate a chat history between a user and an assistant, where:

- The user makes an initial request that requires a tool call.

- The assistant responds by calling the tool and providing the result.

- After a time gap (implied or explicit), the user makes a follow-up request that could \textbf{necessarily use the same kind of information as before but the result from the tool call would be different if made at a different time from the previous tool call}. Avoid asking exactly the same question as before.
\\

To boost the quality and diversity of the samples, please use a variety of strategies for the follow-up question. Here are some examples:

1. \textbf{Repeated After Failure:} A previous tool call resulted in a failure (e.g., limited number of available resources). Later, the user repeats the request or tries a similar one for a different entity, hoping the situation has changed. The final user message should reflect this repeated attempt.

2.  \textbf{User Confirmation:} A previous tool call was successful. Later, the user asks for confirmation of the result of that tool call (e.g., 'Is X still in Y state?', 'Was my request processed?'). This should be the final request.

3.  \textbf{Request Repeat:} A previous tool call was successful. After some intervening turns on other topics, a user repeats the exact same request, as if they have forgotten the previous successful interaction. The repeated request should be the last one.

4.  \textbf{In Context Availability:} The tool provides information about a resource's state that has a limit (e.g., number of available slots, capacity). The user performs actions that change this state until the limit is reached. Unaware of the state, the user then makes another request that would exceed the limit. This should be the last request.\\

Please follow the structure and format of the provided example, including roles, tool calls, and JSON formatting. The last message should be a user message. Do not include any time indicators such as 'still' in the last message. \\

Here is the example:

Topic: [example\_topic]

Description: [example\_description]

Example: [example]\\

Now, create a similar chat trajectory for the following topic.

Topic: [topic]

Description: [description]\\

You can use the following strategy for the follow-up question if it is appropriate: [preferred\_strategy]. Feel free to use other strategies if they are more appropriate.

Output in the same JSON format as above.

\end{tcolorbox}
\caption{\textbf{The prompt used for generating \emph{read+write} samples with an in-context example.} To fully leverage the instruction following capability of the model, we generate the samples for one variant at a time by passing in a \emph{preferred\_strategy}.}
\label{fig:prompt_rw}
\end{figure*}
\begin{table*}[h]
\scriptsize
\centering
\begin{tabular}{p{0.28\textwidth}p{0.43\textwidth}p{0.05\textwidth}p{0.06\textwidth}}
\toprule
\textbf{\small Scenario Name} & \textbf{\small Description} & \textbf{\small I/O} & \textbf{\small Samples} \\
\midrule
\multicolumn{3}{l}{\textbf{\small Time Sensitivity: Low}} & \textbf{544} \\
\midrule
\textbf{Regulatory Information Service} & Provides information on regulations like tax brackets. & \textit{R} & 20 \\
\textbf{University Degree Requirements} & Lists the official curriculum and graduation requirements. & \textit{R} & 21 \\
\textbf{Public Health Screening Guidelines} & Provides public health recommendations for medical screenings. & \textit{R} & 20 \\
\textbf{Airline Baggage Policy} & Current luggage allowance, size limits, and fees for an airline carrier. & \textit{R} & 20 \\
\textbf{Bank Interest Rate Checker} & Retrieves the advertised Annual Percentage Yield (APY) for a bank. & \textit{R} & 20 \\
\textbf{Recorded Property Deed Lookup} & Retrieves public land-record documents for a parcel. & \textit{R} & 20 \\
\textbf{Astronomical Object Info} & Returns static data and properties about celestial bodies like planets. & \textit{R} & 20 \\
\textbf{Company Policy Lookup} & Retrieves corporate policies like WFH rules and benefits. & \textit{R} & 20 \\
\textbf{QRH Guidelines} & Provides quick-reference pilot procedures for common failures. & \textit{R} & 17 \\
\textbf{Protected CITES / Protected Places Rules} & Returns protected-area rules on access, hunting, and activities. & \textit{R} & 18 \\
\textbf{Manufacturer Product Manual Retriever} & Retrieves official manuals and technical specs for a model. & \textit{R} & 20 \\
\textbf{ISO / Industry Standard Text} & Retrieves published text for a specific standard version. & \textit{R} & 20 \\
\textbf{Chemical Safety Data Sheet (SDS)} & Provides hazards, PPE guidance, and safe handling instructions. & \textit{R} & 20 \\
\textbf{Programming Language Syntax Helper} & Shows correct syntax for a specific language/version command. & \textit{R} & 20 \\
\textbf{Public Attraction Hours} & Retrieves typical operating hours that may vary seasonally. & \textit{R} & 20 \\
\textbf{Store Location Finder} & Finds nearest chain store address and basic location details. & \textit{R} & 20 \\
\textbf{Airport Code Lookup} & Provides IATA and ICAO codes for airports worldwide. & \textit{R} & 10 \\
\textbf{Endangered Species List} & Checks conservation status from official lists and registries. & \textit{R} & 20 \\
\textbf{Streaming Service Library Checker} & Checks if a title is currently on a platform. & \textit{R} & 18 \\
\textbf{Office Finder} & Returns staff/professor office location within a building. & \textit{R} & 20 \\
\textbf{Class Room Finder} & Returns a student's classroom location with directions. & \textit{R} & 17 \\
\textbf{Patent Metadata Lookup} & Retrieves bibliographic fields for a published patent record. & \textit{R} & 20 \\
\textbf{Archived Government Legislation Text} & Retrieves the official as-published text of a past law. & \textit{R} & 9 \\
\textbf{UN Country Membership List} & Lists UN members and their official join dates. & \textit{R} & 20 \\
\textbf{EV Station Status} & Reports whether a charging station is working or broken. & \textit{R} & 14 \\
\textbf{Developing/Developed Countries Info} & Provides country development classification and brief context. & \textit{R} & 20 \\
\textbf{Public Transportation Route Mapper} & Describes fixed bus/subway routes and primary stops. & \textit{R} & 20 \\
\textbf{City Population Data} & Retrieves the latest official estimate for a city population. & \textit{R} & 20 \\
\textbf{Pharmacological Database} & Lists drug uses, interactions, contraindications, and properties. & \textit{R} & 20 \\
\midrule
\multicolumn{3}{l}{\textbf{\small Time Sensitivity: Medium}} & \textbf{681} \\
\midrule
\textbf{Job Board Live Listings} & Searches for currently open positions and their application status. & \textit{R} & 20 \\
\textbf{Weather Forecast Service} & Fetches the weather forecast for the next 1-10 days. & \textit{R} & 20 \\
\textbf{Package / Shipment Tracker} & Provides the current transit status and estimated time of arrival. & \textit{R} & 23 \\
\textbf{Laundry Service Order Status} & Checks if clothes dropped off earlier are ready for pickup today. & \textit{R} & 24 \\
\textbf{In-Game Marketplace Price Checker} & Current lowest price for a tradable item in a multiplayer online game. & \textit{R} & 24 \\
\textbf{Airport Security Line Wait Time} & Shows real-time airport security wait times. & \textit{R} & 18 \\
\textbf{Tide Height Tracker} & Shows the live measured water level at a specific coastal point. & \textit{R} & 20 \\
\textbf{Pollution Emission Monitor} & Shows real-time emission data from a monitoring site. & \textit{R} & 20 \\
\textbf{Movie Showtimes Finder} & Retrieves local showtimes for the current week schedule. & \textit{R} & 21 \\
\textbf{Currency Exchange Rate} & Provides daily FX rates for travel or budgeting use. & \textit{R} & 21 \\
\textbf{Grocery Store Sale Checker} & Lists weekly sale items and discounts at a store. & \textit{R} & 21 \\
\textbf{Library Book Availability} & Checks whether a physical book is on shelf or out. & \textit{R} & 19 \\
\textbf{Real Estate Listings} & Shows current homes for sale or rent in an area. & \textit{R} & 30 \\
\textbf{Ski Resort Snow Report} & Reports snow depth, lift status, and open trails today. & \textit{R} & 27 \\
\textbf{Hiking Trail Condition Report} & Summarizes recent trail conditions, closures, and hazards. & \textit{R} & 27 \\
\textbf{Surf Report} & Predicts wave height, winds, and tide timing at a spot. & \textit{R} & 24 \\
\textbf{Visa Case Status} & Checks government case stage updates and processing status. & \textit{R} & 21 \\
\textbf{Subscription / Account Status} & Checks if a subscription is active and its expiry time. & \textit{R} & 21 \\
\textbf{Hotel Room Availability \& Rates} & Checks for room availability and fluctuating prices at a hotel. & \textit{R\&W} & 39 \\
\textbf{Tee Time Booker} & Finds and books available times to play at a golf course. & \textit{R\&W} & 41 \\
\textbf{Prescription Fulfillment \& Hold} & Checks stock and places a hold on a prescription for immediate pickup. & \textit{R\&W} & 40 \\
\textbf{Flight Search \& Booking} & Live flight search, and the ability to hold or book a flight. & \textit{R\&W} & 40 \\
\textbf{Appointment Availability (Clinic)} & Finds open clinic slots for non-emergency appointments nearby. & \textit{R\&W} & 40 \\
\textbf{Appointment Availability (Haircut)} & Finds open salon/barber slots and supports scheduling. & \textit{R\&W} & 40 \\
\textbf{Rental Car Availability \& Rate Quote} & Checks rental inventory and returns a quoted daily rate. & \textit{R\&W} & 40 \\
\midrule
\multicolumn{3}{l}{\textbf{\small Time Sensitivity: High}} & \textbf{639} \\
\midrule
\textbf{Live Sports Betting Odds Retrieval} & Provides rapidly fluctuating betting odds for an in-progress game. & \textit{R} & 20 \\
\textbf{Live Auction Price Retrieval} & Checking price in a fast-paced online auction. & \textit{R} & 20 \\
\textbf{E-commerce Product Stock Checker} & Checks current price and inventory for a retail product. & \textit{R} & 35 \\
\textbf{Live Traffic Navigation \& ETA} & Calculates travel ETAs using real-time traffic data. & \textit{R} & 24 \\
\textbf{Live Sports Score Tracker} & Retrieves the current score and status of an ongoing sports game. & \textit{R} & 21 \\
\textbf{Live Vehicle GPS Tracker} & Provides the precise location of a vehicle like a delivery truck. & \textit{R} & 22 \\
\textbf{Live Weather Sensor Data} & Reads data directly from a weather sensor (e.g., wind speed). & \textit{R} & 24 \\
\textbf{Train Delay Tracker} & Reports active delays and disruptions for train services. & \textit{R} & 20 \\
\textbf{Restaurant Wait Time Checker} & Gives the current estimated wait time for a table at a restaurant. & \textit{R} & 19 \\
\textbf{Stock Market Order Book} & Shows the real-time list of buy and sell orders for a particular stock. & \textit{R} & 20 \\
\textbf{Cryptocurrency Exchange Retrieval} & Get cryptocurrency live trade prices. & \textit{R} & 18 \\
\textbf{Cloud Server Load Balancer} & Routes processes based on real-time server CPU load. & \textit{R\&W} & 40 \\
\textbf{ICU Vitals Monitor} & Streams live patient vitals and controls alarm thresholds. & \textit{R\&W} & 40 \\
\textbf{Power Grid Control System} & Manages power grid distribution with real-time control signals. & \textit{R\&W} & 40 \\
\textbf{Ride-Hailing Dispatch} & Requests immediate rideshare pickup and returns live ETA. & \textit{R\&W} & 25 \\
\textbf{Taxi Dispatch System} & Finds available taxis nearby and dispatches a pickup. & \textit{R\&W} & 40 \\
\textbf{Emergency Alert / 911 Dispatch Interface} & Submits emergency details and receives responder status updates. & \textit{R\&W} & 18 \\
\textbf{Ambulance / Emergency Resource Dispatch} & Allocates emergency vehicles and tracks dispatch progress live. & \textit{R\&W} & 40 \\
\textbf{Urban Parking-Spot Reservation} & Claims a scarce spot that can be taken instantly. & \textit{R\&W} & 82 \\
\textbf{Concert Ticket} & Checks live ticket availability and completes a purchase. & \textit{R\&W} & 21 \\
\textbf{Gasoline Dispatcher} & Dispatches fuel, tracks volume remaining, and fulfillment status. & \textit{R\&W} & 29 \\
\textbf{Stock Trading (Order Submission)} & Submits a trade order for immediate market execution. & \textit{R\&W} & 21 \\
\bottomrule

\end{tabular}
\vspace{-8pt}
\caption{\textbf{Complete inventory of scenarios included in \dsname.} Scenarios are categorized into three levels of time sensitivity based on their temporal dynamics, with brief descriptions provided. We also indicate whether each scenario is \emph{read-only} (\textit{R}) or \emph{read+write} (\textit{R\&W}).}
\label{tab:scenario_invetory}
\end{table*}

\begin{figure*}[!h]
\begin{tcolorbox}[
  enhanced, 
  colframe=orange!75!black, 
  colback=white, 
  coltitle=white, 
  colbacktitle=orange!75!black, 
  width=\linewidth, 
  arc=2mm, 
  auto outer arc, 
  boxrule=0.5pt, 
  left=10pt, 
  right=10pt, 
  drop shadow={black!50!white},
  top=10pt, 
  bottom=10pt, 
  title=\textbf{Prompt Template for Sample verification}, 
  fonttitle=\bfseries, 
  title code={\node[rounded corners, fill=blue!75!black, draw=none, text=white] at (frame.title) {\textbf{xxx}};}, 
  attach boxed title to top center={yshift=-2mm}, 
  boxed title style={sharp corners, size=small}, 
]
You are an expert evaluator for AI-generated chat trajectories. Your task is to determine if a given sample meets a specific criterion for evaluating an agent's temporal awareness.\\

Here is the sample to evaluate:\\
Sample: [sample]\\

Please evaluate the sample based on the following criterion:\\
1. \textbf{No Explicit Hints:} The final user question must not contain explicit hints for calling a tool, such as "try again now".\\
2. \textbf{No Missing Premise:} With the help of rounds of tool calls at the moment, the agent must have all information needed to answer the user's final message. Be strict on this criterion.\\
3. \textbf{Temporal Dependency:} The agent would be wrong if it ignores the time elapse between the final user message and prior messages when making a decision whether or not to call a tool. We want the agent to be punished for ignoring the time gap and blindly rely on previous tool call result, or for excessively repeating a tool call that is not necessary since the time gap is small or environment is static.\\

Provide your evaluation in a JSON format with two keys: 'is\_valid' (boolean) and 'reason' (a string briefly explaining your decision).
\end{tcolorbox}
\caption{The prompt given to \texttt{GPT-4.1} to perform first round quality filtering.}
\label{fig:verify}
\end{figure*}

\begin{figure*}[!h]
\begin{tcolorbox}[
  enhanced, 
  colframe=red!75!black, 
  colback=white, 
  coltitle=white, 
  colbacktitle=red!75!black, 
  width=\linewidth, 
  arc=2mm, 
  auto outer arc, 
  boxrule=0.5pt, 
  left=10pt, 
  right=10pt, 
  drop shadow={black!50!white},
  top=10pt, 
  bottom=10pt, 
  title=\textbf{Prompting-based alignment with few-shot example rules}, 
  fonttitle=\bfseries, 
  title code={\node[rounded corners, fill=blue!75!black, draw=none, text=white] at (frame.title) {\textbf{xxx}};}, 
  attach boxed title to top center={yshift=-2mm}, 
  boxed title style={sharp corners, size=small}, 
]
Note that the environment may be dynamic. Be aware of the time elapsed. Depending on the scenario and how much time has passed, whether it's preferable to call a tool or not can vary.\\ 

For example, suppose you were a smart garden assistant with access to a tool that measures soil moisture levels for houseplants. If you successfully retrieved the moisture level just five minutes ago and it read ``Wet'', it is physically impossible for the soil to have dried out in such a short time. Therefore, calling the tool again is unnecessary. However, if four days have passed since the last check, natural evaporation has certainly occurred, so you must call the tool again to see if the plants need watering.\\

As another example, suppose you were a student aid agent checking an online gradebook to see if a professor has posted final exam results. If you checked the portal a few seconds ago and the grade was ``Pending'', calling the tool again immediately is not needed, as human grading takes time. But if a week has passed since your last check, it is highly probable the professor has finished grading, so it is clearly better to query the tool again to get the latest result.
\end{tcolorbox}
\caption{\textbf{Prompt used to explicitly align models' tool-use decisions with human expectations.} Few-shot example rules are used to illustrate when tool calls are appropriate or unnecessary depending on elapsed time, providing models with explicit guidance. \textbf{Note that the scenarios mentioned in the prompt do not overlap with our coverage in \dsname.}}
\vspace{-5pt}
\label{fig:rule}
\end{figure*}

\onecolumn

\end{document}